\documentclass[10pt,twocolumn,letterpaper]{article}

\usepackage{cvpr}              
\definecolor{cvprblue}{rgb}{0.21,0.49,0.74}
\usepackage[pagebackref,breaklinks,colorlinks,allcolors=cvprblue]{hyperref}
\usepackage[normalem]{ulem}
\usepackage{enumitem} 
\usepackage{algorithm}
\usepackage{algpseudocode}
\algrenewcommand\algorithmicrequire{\textbf{Require:}}
\algrenewcommand\algorithmicensure{\textbf{Ensure:}}
\usepackage{hyperref}

\usepackage{amsmath,amssymb,amsthm}
\usepackage{amsmath,amssymb}
\usepackage{mathrsfs}          
\usepackage{caption}

\usepackage[table]{xcolor}
\definecolor{AvgRow}{HTML}{F2F2FF}
\usepackage{mathtools}
\usepackage{placeins}
\usepackage{array}
\usepackage[accsupp]{axessibility}  
\usepackage{hyphenat}

\algrenewcommand\algorithmicrequire{\textbf{Input:}}
\algrenewcommand\algorithmicensure{\textbf{Output:}}
\usepackage{xspace}
\newcommand{\method}{O2MAG\xspace}
\usepackage{booktabs,multirow,pifont}
\newcommand{\cmark}{\ding{51}} 

\usepackage{relsize}

\def\paperID{} 
\def\confName{CVPR}
\def\confYear{2026}

\title{\texorpdfstring{%
One-to-More: High-Fidelity Training-Free Anomaly Generation \\ with Attention Control \vspace{-0.35cm}%
}{One-to-More: Tri-branch Attention Grafting for Training-Free Anomaly Generation}}

\author{
    Haoxiang Rao\textsuperscript{1}, 
    Zhao Wang\textsuperscript{2}, 
    Chenyang Si\textsuperscript{1}, 
    Yan LYU\textsuperscript{2}, 
    Yuanyi Duan\textsuperscript{3}, 
    Fang Zhao\textsuperscript{1,$\dagger$}, 
    Caifeng Shan\textsuperscript{1,$\dagger$}
    \\
    \textsuperscript{1}Nanjing University, 
    \textsuperscript{2}China Mobile Zijin Innovation Institute \\
    \textsuperscript{3}Shandong University of Science and Technology \\
    {\tt\small echrao@smail.nju.edu.cn, \{wangzh8, lvyansgs\}@js.chinamobile.com} \\
    {\tt\small \{chenyang.si.mail, duanyuanyi46, zhaofang0627, caifeng.shan\}@gmail.com}
}

\begin{document}

\twocolumn[{
\renewcommand\twocolumn[1][]{#1}
\maketitle
\vspace{-0.9cm}
\begin{center}
    \centering
    \includegraphics[page=1,width=\linewidth]{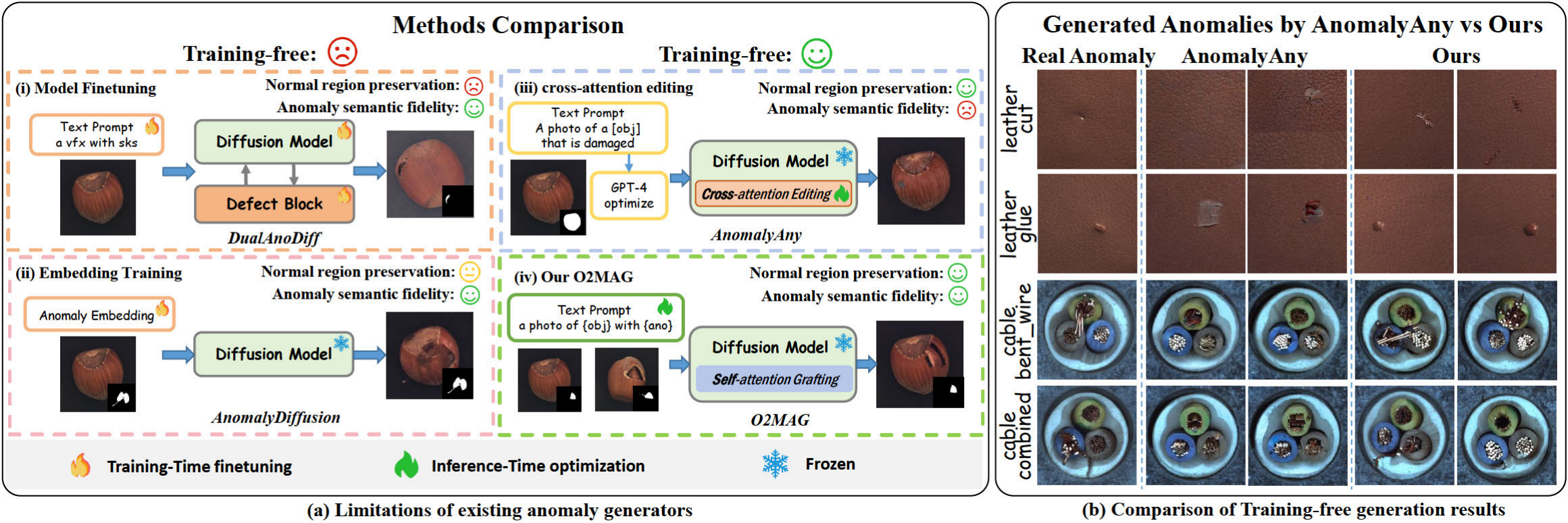}
    \\[-1.2ex]
    \captionof{figure}{\textbf{Left}: Comparison of diffusion-based anomaly generation. Training-based approaches either (i) add defect block to learn the anomaly distribution or (ii) train embeddings by textual-inversion to mimic anomalous visual styles; whereas, (iii) existing training-free method, AnomalyAny, fails to express precise and realistic anomaly semantics while (iv) our proposed training-free O2MAG delivers background-faithful synthesis with diverse, localized anomalies. \textbf{Right}: Comparison of training-free methods between AnomalyAny and our O2MAG. In AnomalyAny, the glue-on-leather and wire-in-cable deviate from the real defect distribution.}
    \label{fig:P1}
\end{center}
}]

\begingroup
\renewcommand\thefootnote{}
\footnotetext{\textsuperscript{$\dagger$}Corresponding authors.}
\endgroup

\begin{abstract}
Industrial anomaly detection (AD) is characterized by an abundance of normal images but a scarcity of anomalous ones. Although numerous few-shot anomaly synthesis methods have been proposed to augment anomalous data for downstream AD tasks, most existing approaches require time-consuming training and struggle to learn distributions that are faithful to real anomalies, thereby restricting the efficacy of AD models trained on such data. To address these limitations, we propose a training-free few-shot anomaly generation method, namely \textbf{\method}, which leverages the self-attention in \textbf{One} reference anomalous image \textbf{to} synthesize \textbf{More} realistic anomalies, supporting effective downstream anomaly detection. Specifically, \method manipulates three parallel diffusion processes via self-attention grafting and incorporates the anomaly mask to mitigate foreground-background query confusion, synthesizing text-guided anomalies that closely adhere to real anomalous distributions. To bridge the semantic gap between the encoded anomaly text prompts and the true anomaly semantics, Anomaly-Guided Optimization is further introduced to align the synthesis process with the target anomalous distribution, steering the generation toward realistic and text-consistent anomalies. Moreover, to mitigate faint anomaly synthesis inside anomaly masks, Dual-Attention Enhancement is adopted during generation to reinforce both self- and cross-attention on masked regions. Extensive experiments validate the effectiveness of \method, demonstrating its superior performance over prior state-of-the-art methods on downstream AD tasks.
\vspace{-0.5cm}

\end{abstract}

\section{Introduction}
\label{sec:intro}
In industrial manufacture, visual anomaly inspection, encompassing detection, localization, and classification, underpins quality control and yield improvement~\cite{lin2024comprehensive,huang2025deep}. However, real-world industrial production commonly exhibits data imbalance, i.e., abundant normal images and scarce anomalous ones. While training solely on normal data can identify anomalies deviating from the normal distribution~\cite{yao2023explicit,li2024promptad}, precise anomaly localization and classification still require a sufficient number of representative, labeled anomaly examples. Consequently, recent studies have exploited generative models to synthesize realistic anomalous images to enhance supervised anomaly inspection~\cite{duan2023few,hu2024anomalydiffusion}.

Diffusion models~\cite{rombach2022high}, with stable iterative denoising and strong conditioning, have increasingly supplanted GANs~\cite{goodfellow2020generative} as the backbone for anomaly generation~\cite{dai2024generating,jin2025dual,dai2025seasfewshotindustrialanomaly,choi2025magic,zhang2024realnet}. In the diffusion setting, training-based approaches fall into two categories: model finetuning (\cref{fig:P1}(a.i)) and embedding training (\cref{fig:P1}(a.ii)). Model-finetuning methods are frequently based on DreamBooth~\cite{ruiz2023dreambooth}, adapting the backbone to bind a rare identifier token to the anomaly concept while preserving key visual attributes~\cite{ali2024anomalycontrol,dai2025seasfewshotindustrialanomaly,jin2025dual,song2025defectfill} and many also attach a defect block to explicitly model the anomalous foreground (e.g., DualAnoDiff~\cite{jin2025dual}, DefectDiffu~\cite{song2025defectfill}). Although this can produce semantically diverse anomalous images, normal regions are often not preserved, which substantially reduces the realism of the synthesized images. Embedding-training methods freeze the diffusion backbone and learn anomaly embeddings via textual inversion~\cite{hu2024anomalyxfusion,hu2024anomalydiffusion} to represent specific anomaly concepts~\cite{gal2022image}. 

However, the foregoing methods demand substantial training compute and storage overhead, motivating training-free alternatives. AnomalyAny~\cite{sun2025unseen} (\cref{fig:P1}(a.iii)) refine cross-attention to attend anomaly tokens and strengthen their activations to realize generating anomaly concepts. However, as shown in \cref{fig:P1}(b), it struggles to produce realistic anomaly appearances, offers limited control over anomaly types and spatial layouts, and cannot produce pixel-accurate masks, thereby limiting downstream anomaly detection and classification. To leverage abundant global-layout and texture information in self-attention, we introduce \method (\cref{fig:P1}(a.iv)), which performs training-free synthesis via text-embedding editing and attention modulation, producing realistic anomalies in specified regions while preserving background fidelity, substantially improving downstream anomaly detection accuracy.


\method comprises three modules: the main Tri-branch Attention Grafting (TriAG) for anomalous features transfer, Anomaly-Guided Optimization (AGO) to refine anomaly text embeddings, and Dual-Attention Enhancement (DAE) to ensure complete filling within the target mask. TriAG is the backbone with three diffusion branches: reference-anomaly, normal-image, and target-anomaly. Within the target mask, self-attention grafting module enables cross-branch information exchange, which queries foreground anomaly features from the reference branch and background normal features from the normal branch, producing diverse anomalous foregrounds while preserving normal backgrounds without any model fine-tuning. Because anomalies are scarce in SD training and simple prompts attend LAION-5B~\cite{schuhmann2022laion} biases misaligned with industrial defects, AGO keeps the pipeline training-free by a lightweight optimization of the anomaly text embedding via a reconstruction loss to a reference anomaly image. Finally, to prevent weak or incomplete anomalies inside the mask, DAE reweights self- and cross-attention toward defect regions at selected timesteps, improving small-defect fidelity and mask-consistent synthesis. To quantitatively assess our method, we evaluate on MVTec-AD dataset~\cite{bergmann2019mvtec} and improves pixel-level AP by \(+1.8\%\) and F1-max by \(+2.0\%\) and raises classification accuracy by \(+12.1\%\). In summary, the key contribution of our approach lies in:
\begin{itemize}
  \item We propose a training-free, few-shot tri-branch diffusion framework that synthesizes mask-localized, background-faithful anomalies by grafting self-attention from reference anomaly and normal branch into a target branch.
  \item We perform inference-time optimization of the anomaly text embedding via reconstruction loss on a reference anomaly image, yielding domain-aligned text conditioning without fine-tuning.
  \item During denoising, Dual-Attention Enhancement is injected at specific timesteps to upweight self-/cross-attention within the mask region, improving the alignment between synthesized anomalies and the mask.
\end{itemize}

\section{Related work}
\label{sec:related_works}
{%
\setlength{\belowcaptionskip}{-10pt} 
\begin{figure*}[t]
  \centering
  \includegraphics[width=\textwidth]{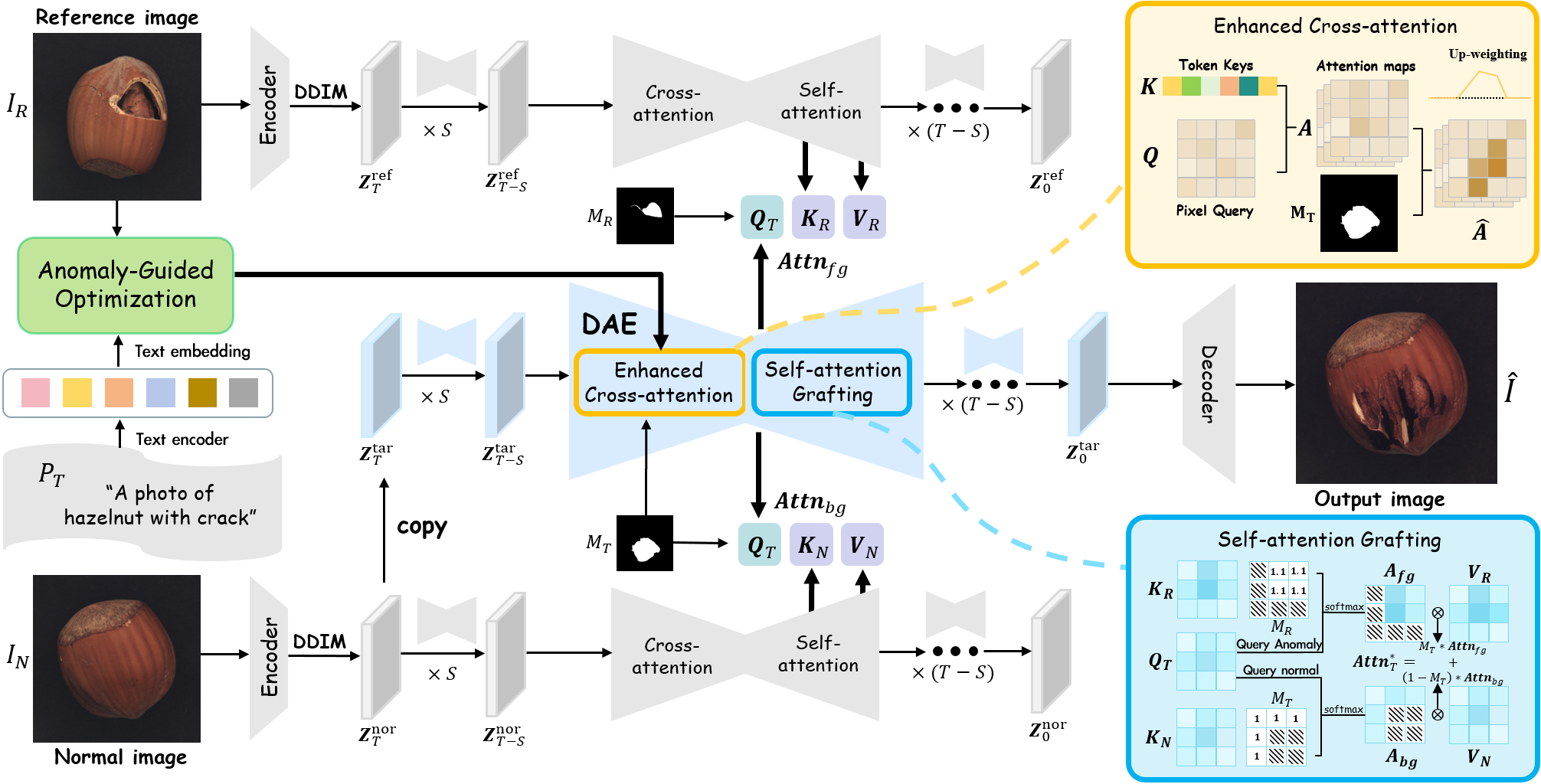}
  \caption{Overview of the proposed \method. Our method synthesize anomalies by coordinating self-attention in three parallel diffusion process. DAE operates on the target branch’s self-/cross-attention to enforce full-mask filling, while AGO refines the anomaly text embedding to align semantics with reference anomaly image to make generation more realistic.}
  \label{fig2:method}
\end{figure*}
}
\subsection{Anomaly generation}
Early anomaly synthesis relied on hand-crafted data augmentation to synthesize pseudo-anomalies. CutPaste~\cite{li2021cutpaste} and DRAEM~\cite{zavrtanik2021draem} paste image patches or unrelated textures onto normal images rather than modeling the full image distribution, generating geometrically or photometrically inconsistent anomalies that limit realism and diversity.

\textbf{GAN-based generation.} GANs~\cite{goodfellow2020generative} were subsequently adopted for industrial anomaly generation. Methods such as SDGAN~\cite{niu2020defect} and DefectGAN~\cite{zhang2021defect} learn mappings between normal and anomaly domains to synthesize defects. DFMGAN~\cite{duan2023few} employs a two-stage scheme to model the distributions of normal and anomalous images separately. However, under anomaly-scarce regimes, GANs suffer from mode collapse and gradient instability~\cite{mescheder2018training,duan2023few}, reducing diversity.

\textbf{Diffusion-based generation.} 
Diffusion models are increasingly adopted in anomaly synthesis due to their strong generalization. AnomalyDiffusion~\cite{hu2024anomalydiffusion} and AnoGen~\cite{gui2025fewshotanomalydrivengenerationanomaly} freeze the diffusion backbone and learn anomaly embeddings that encode specific anomaly concepts~\cite{gal2022image}, but they often miss fine-grained defect structure. Recent works have shifted toward DreamBooth-based finetuning that fine-tunes the entire backbone to bind a rare identifier token to anomaly concept~\cite{ruiz2023dreambooth}. DualAnoDiff~\cite{jin2025dual} couples a global image branch with a defect-patch branch to produce anomalies. DefectFill~\cite{song2025defectfill} fine-tunes an inpainting diffusion model to associate foreground anomalies with rare identifiers. SeaS~\cite{dai2025seasfewshotindustrialanomaly} further introduces an unbalanced abnormal text prompt to model the distributional discrepancy between normal products and diverse anomalies. However, under few-shot settings, these methods struggle to faithfully model the anomaly data distribution while avoiding overfitting. Despite incurring computational cost and long training time, the trained generators still deviate markedly from the real-data distribution.

\subsection{Text-to-Image Editing and Synthesis}

Image editing refers to introducing user-provided conditions (e.g., image or text) into generative process to constrain the output’s structure and semantics. Textual Inversion~\cite{gal2022image} and DreamBooth~\cite{ruiz2023dreambooth} bind user-image features to rare identifier tokens for personalized generation, but demand substantial per-object fine-tuning and afford weak control over layout and background. P2P~\cite{hertz2022prompt} and Attend-and-Excite~\cite{chefer2023attend} reveal rich semantics encoded in cross-attention as the correspondence between text tokens and spatial layout. However, because cross-attention is formed by matching spatial features to word embeddings, it captures only coarse object-level regions and cannot expressed local spatial details that are not explicitly specified in the original prompt. Complementary approaches—including MasaCtrl~\cite{cao2023masactrl}, PnP~\cite{tumanyan2023plug}, and MCA-Ctrl~\cite{yang2025multi} edit self-attention to constrain global structure and local deformations, enabling more complex non-rigid edits.

Despite the rich semantics encoded in attention, training-free attention modulation for industrial anomaly synthesis remains underexplored. Training-free methods, TF\textsuperscript{2}~\cite{yu2024tf2} and AnomalyAny~\cite{sun2025unseen}, lacks fine-grained spatial and semantic control. To address above issues, we propose an one-shot anomaly-guided training-free approach enables fine-grained and more realistic anomaly synthesis by manipulating attention features to steer generation, avoiding finetuning and large checkpoint storage.
\section{Preliminary}
\subsection{Stable Diffusion}
Stable Diffusion~\cite{rombach2022high} employs a strategy based on latent space, where the VAE is trained to map images $x\in\mathbb{R}^{H\times W\times3}$ in RGB space to low-dimensional latent representations  $\boldsymbol{z}=\mathcal{E}(x)$, where $\boldsymbol{z}\in\mathbb{R}^{h\times w\times c}$. The process of adding and removing Gaussian noise is carried out in this latent space, with the final denoising output being mapped back to the pixel space by the decoder $\tilde{x}=\mathcal{D}(\boldsymbol{z})=\mathcal{D}(\mathcal{E}(x))$. The model includes a CLIP text encoder~\cite{radford2021learning} $\tau_\theta$ that projects the text condition $y$ to an intermediate representation $\tau_\theta(y)\in\mathbb{R}^{m\times d_\tau}$, and the conditional Stable Diffusion can be learned via
\begin{equation}
L_{SD}:=\mathbb{E}_{\mathcal{E}(x),y,\epsilon\sim\mathcal{N}(0,1),t}\left[\|\epsilon-\epsilon_{\theta}(\boldsymbol{z}_{t},t,\tau_{\theta}(y))\|_{2}^{2}\right].
\end{equation}
where $\boldsymbol{z}_t$ denotes the noised latent at diffusion step $t$.

\subsection{Attention Mechanism in Stable Diffusion}
\label{sec:Attention_Mechanism}
Each U-Net block~\cite{ronneberger2015u} in the SD model includes a residual block~\cite{he2016deep}, a self-attention module, and a cross-attention module~\cite{vaswani2017attention}. Text guidance is injected via cross-attention, while self-attention integrates global context to preserve structure and boundaries. Both are applied to intermediate feature maps at \(64\times64\), \(32\times32\), \(16\times16\), and \(8\times8\) resolutions. The attention map is computed as
\begin{equation}
\boldsymbol{A}=\mathrm{softmax}\!\left(\frac{\boldsymbol{Q}\boldsymbol{K}^{\mathrm T}}{\sqrt{d_k}}\right),
\end{equation}
The matrix $\boldsymbol{A}$ represents the similarity between queries $\boldsymbol{Q}$ and keys $\boldsymbol{K}$. $\boldsymbol{A}$ is then used to aggregate the values $\boldsymbol{V}$ to produce the attention output \(\mathrm{\boldsymbol{Attn}}(\boldsymbol{Q},\boldsymbol{K},\boldsymbol{V})=\boldsymbol{A}\,\boldsymbol{V}\), where $\boldsymbol{Q}$ denotes queries projected from spatial features, and $\boldsymbol{K},\boldsymbol{V}$ are projected from spatial features (self-attention) or text embeddings (cross-attention) via the corresponding projection matrices. The attention output is subsequently used to update the spatial features.


Prior studies have shown that SD attention layers capture global layout and content-formation signals. Cross-attention aligns spatial pixels with text embeddings; self-attention features serve as plug-and-play features that can be injected into specific U-Net layers to realize image transformations. Accordingly, we manipulate attention features to enable efficient anomaly image generation.
\section{Methodology}

We propose \method, a training-free, few-shot diffusion framework for anomaly generation that exploits intrinsic priors of diffusion models. With only one single reference anomaly image, \method can synthesize more diverse and realistic anomaly samples. As illustrated in \cref{fig2:method}, \method comprises three parallel diffusion processes. Through self-attention feature grafting, it edits anomalous appearance while preserving the normal background (\autoref{sec:TriAG}). In addition, Anomaly-Guided Optimization Module (\autoref{sec:AGO}) aligns the text prompt embeddings with the target anomaly semantics, injecting richer defect semantics to improve generation quality. Besides, by upweighting cross- and self-attention to boost responses within the anomaly region, ensuring full filling of the target mask (\autoref{sec:DAE}).


\subsection{Tri-branch Attention Grafting}
\label{sec:TriAG}
{%
\setlength{\belowcaptionskip}{-10pt} 
\begin{figure}[t]
  \centering
  \includegraphics[width=\linewidth]{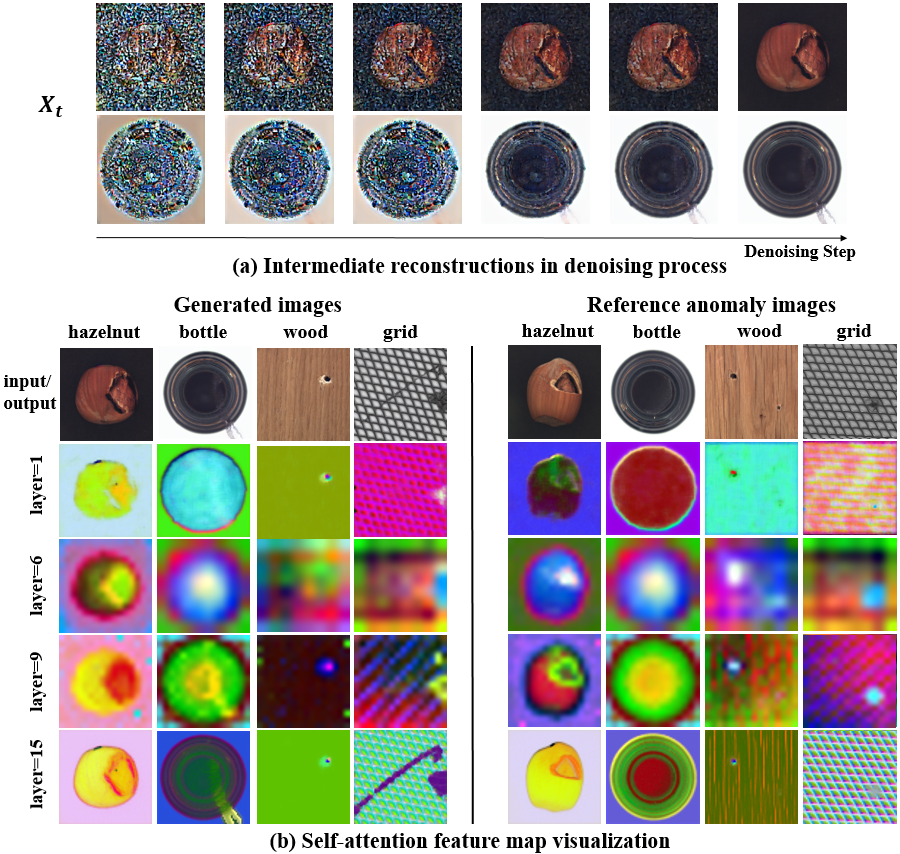} 
  \caption{(a) The intermediate reconstruction during the iterative denoising process, and (b) visualization of the self-attention map A at the 30th sampling step. Principal components preserve the image layout and regions rendered with the same color share semantic attributes.}
  \label{fig3:self-attention}
\end{figure}
}

In our work, we leverage SD~\cite{rombach2022high} to generate more visual anomalies. Given one reference anomaly image–mask pair \((I_{\mathrm{R}}, M_{\mathrm{R}})\), a normal image \(I_{\mathrm{N}}\), and a target anomaly mask \(M_{\mathrm{T}}\), our goal is to synthesize more anomaly images \(\hat{I}\) that conforms to a given text prompt, preserves the background appearance of \(I_{\mathrm{N}}\) outside \(M_{\mathrm{T}}\), and transfers anomaly attribute from \(I_{\mathrm{R}}\) within the region specified by \(M_{\mathrm{T}}\). 

Self-attention constructs attention maps $\boldsymbol{A}$ from the similarities of query–key projections, encoding pairwise spatial affinities across locations. To further explore the established self-similarity, we apply principal component analysis (PCA) to the self-attention maps of anomalous images and visualize the top three components, where regions with similar structure appear in similar colors. The visualization reveals a clear separation between anomalous foreground and normal background in the attention space (see \cref{fig3:self-attention}(b)). This separation, together with prior evidence that tuning-free control of plug-and-play self-attention features enables consistent, structure-preserving synthesis and editing, motivates our design. We keep the current query features \(\boldsymbol{Q}\) unchanged and replace the self-attention keys \(\boldsymbol{K}\) and values \(\boldsymbol{V}\) with those obtained from the reference anomaly, which encode anomalous content and texture to modify the original semantic layout and steer the masked region \(M_{\mathrm T}\) toward the anomaly semantics.


As illustrated in \cref{fig2:method}, TriAG comprises three diffusion processes: the reference-anomaly, normal-image and target-anomaly diffusion branches. Using DDIM inversion~\cite{song2020denoising}, the reference anomaly image \(I_{\mathrm{R}}\) and the normal image \(I_{\mathrm{N}}\) are mapped to initial noise maps \(\boldsymbol{Z}_{T}^{\mathrm{ref}}\) and $\boldsymbol{Z}_{T}^{\mathrm{nor}}$. We initialize the target-anomaly diffusion branch with the normal noise map, \(\boldsymbol{Z}_{T}^{\mathrm{tar}} = \boldsymbol{Z}_{T}^{\mathrm{nor}}\), so that normal content is inherited. Since initialization alone cannot preserve background fidelity, we retain the normal-image branch as a companion to supply background priors. Unlike MasaCtrl~\cite{cao2023masactrl}, images in anomaly datasets are object-centric, and object structure emerges early during denoising (\cref{fig3:self-attention}(a)). Consequently, we start attention grafting at an early denoising timestep \(T_S=5\). Moreover, as shown in \cref{fig3:self-attention}(b), self-attention in the U-Net encoder cannot yet capture a clear layout aligned with the modified prompt. Hence we edit attention at decoder part of self-attention layers \(L_S\in\{9,\dots,16\}\).

Concretely, at the denoising step $t$ and layer $l$, queries from the target-anomaly branch $\boldsymbol{Q}_{T,t,l}$ attend to keys from the reference-anomaly and the normal-image diffusion branches in Eq.~\eqref{eq:triag-attn-first}. To decouple anomalous foreground from background and strictly confine the target diffusion branch to synthesize anomalies only in target region, we mask the attention matrices with \(M_{\mathrm{R}}\) and \(M_{\mathrm{T}}\): the target image is allowed to query only foreground-anomaly features from the reference branch (Eq.~\eqref{eq:fg}), whereas the normal branch contributes only normal background content (Eq.~\eqref{eq:bg}). Finally, the two attention results are gated and fused by \(M_T\) according to Eq.~\eqref{eq:fuse}, yielding the final attention output. This masking localizes anomaly grafting to the target region and enables controllable, spatially precise anomaly synthesis.
\begin{gather}
\boldsymbol{A}_{fg,t,l}=\frac{\boldsymbol{Q}_{T,t,l}\boldsymbol{K}_{R,l}^{\mathrm T}}{\sqrt{d}},\quad
\boldsymbol{A}_{bg,t,l}=\frac{\boldsymbol{Q}_{T,t,l}\boldsymbol{K}_{N,t,l}^{\mathrm T}}{\sqrt{d}} \label{eq:triag-attn-first} \\
\mathrm{\boldsymbol{Attn}}_{\mathrm{fg}}
=\mathrm{softmax}\!\big(\boldsymbol{A}_{fg,t,l}\odot \mathcal{MF}(M_R{=}0,-\infty)\big)\,\boldsymbol{V}_{R,t,l} \label{eq:fg} \\
\mathrm{\boldsymbol{Attn}}_{\mathrm{bg}}
=\mathrm{softmax}\!\big(\boldsymbol{A}_{bg,t,l}\odot \mathcal{MF}(M_T{=}1,-\infty)\big)\,\boldsymbol{V}_{N,t,l} \label{eq:bg} \\
\mathrm{\boldsymbol{Attn}}^{*}_{T,t,l} = M_T \odot \mathrm{\boldsymbol{Attn}_{fg}} + (1 - M_T)\odot \mathrm{\boldsymbol{Attn}_{bg}} \label{eq:fuse}
\end{gather}
where \(d\) is the head dimension, \(\odot\) denotes element-wise multiplication.
\(\mathcal{MF}(\cdot)\) is maskfilling operator.

\subsection{Anomaly-Guided Optimization}
\label{sec:AGO}

We adopt a simple text prompt template, “\textit{A photo of a [cls] with a [anomaly\_type]}”, to specify the object class and anomaly type. However, due to the distribution discrepancy between SD training data and industrial anomaly datasets, the model lacks the capacity to faithfully model real data appearance and semantic relation between object and anomaly attributes. As a result, using the simple text alone to guide generation yields outputs like diamond one in Fig.~\ref{fig:ago}: the normal appearance is not consistent with the distribution of real anomaly data, and the intended anomaly is not conveyed.

{%
\setlength{\textfloatsep}{8pt}      
\setlength{\abovecaptionskip}{9.5pt}  
\setlength{\belowcaptionskip}{-9pt} 
\begin{figure}[t]
  \centering
  \includegraphics[width=\linewidth]{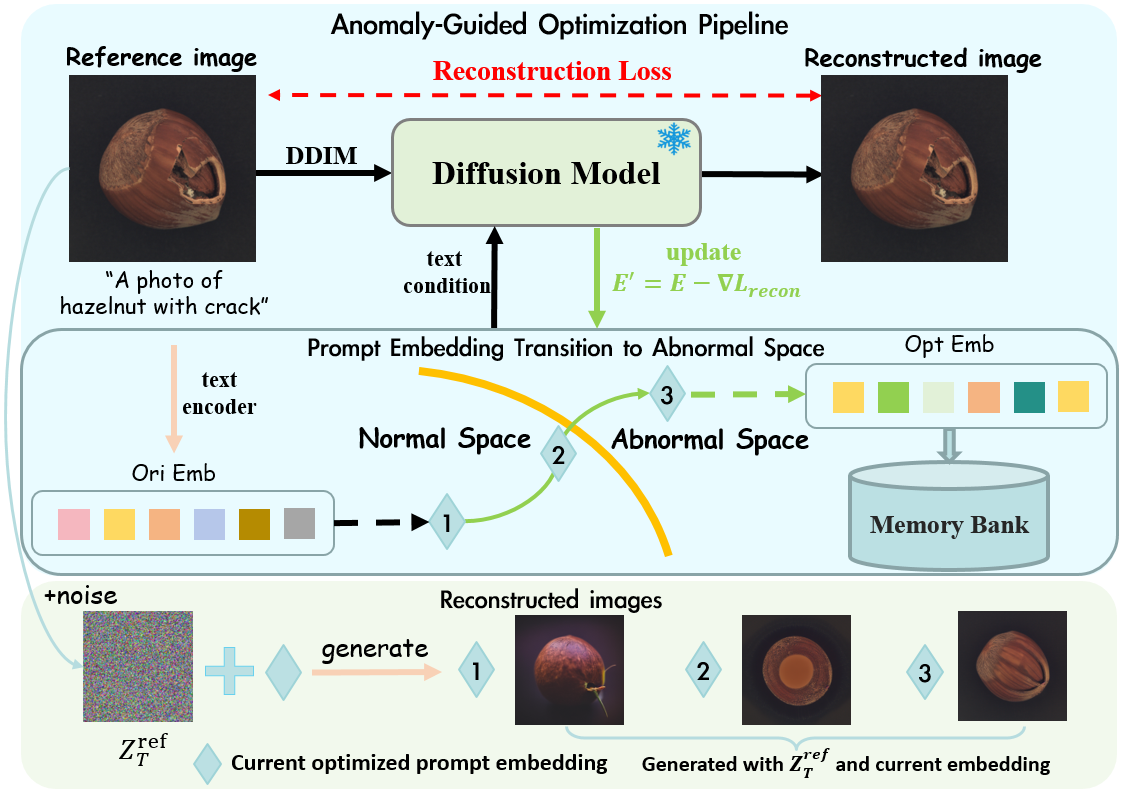} 
  \caption{Anomaly-Guided Optimization (AGO) pipeline. Numbers indicate different synthesized samples from $\boldsymbol{Z}_T^{ref}$ using the current optimized prompt embedding.}
  \label{fig:ago}
\end{figure}
}

To bridge this gap, we propose a lightweight one-shot, reference-anomaly guided prompt optimization that maps normal text semantics into the anomaly space. The target text $y$ is first encoded by a frozen CLIP text encoder~\cite{radford2021learning} $\tau_\theta$ to obtain the original embedding
$\mathbf{\boldsymbol{e}}_{\mathrm{ori}}=\tau_\theta(y)\in\mathbb{R}^{m\times d_\tau},$
where $m$ is the number of tokens and $d_\tau$ is the token-embedding dimension. Then, the DDIM-inverted noise map of the reference anomaly and the text embedding are fed into the frozen SD. We optimize the prompt embedding by minimizing the reconstruction loss in Eq.~\eqref{eq:recon-loss}, thereby pushing it away from normal semantics toward the anomaly space. The optimized embedding can preserve object appearance but encode the intended anomaly attributes.
\begin{equation}\label{eq:recon-loss}
\mathbf{\boldsymbol{e}}^\star
= \arg\min_{\mathbf{\boldsymbol{e}}}\ \mathbb{E}_{t,\boldsymbol{\epsilon}}
\Big[\big\|\boldsymbol{\epsilon}-\epsilon_{\theta}(\mathbf{x}_t,\,t,\,\mathbf{\boldsymbol{e}})\big\|_2^2\Big].
\end{equation}
We further observe that the choice of optimization hyperparameters (e.g., optimization steps and learning rate) substantially affects the quality of the optimized text embedding. In Fig.~\ref{fig:ago}, we show images generated using the inverted noise map of the reference image with the (diamond-labeled) prompt embedding at different optimization steps. Using the original or partially optimized embeddings fails to produce the expected object appearance and anomaly. The exact parameters settings are reported in Sec.~\ref{sec:settings}. By fusing the final optimized prompt embedding (text-level anomaly semantics) with TriAG's image-level anomaly information, our method improves anomaly realism and provides precise, controllable spatial layouts.

\textbf{Negative prompts.}
We additionally adopt negative prompts~\cite{liu2022compositional} to bias synthesis away from normal appearance. 
Let $y$ denote the positive prompt and $y_n$ the negative prompt (e.g., “intact shell,” “no crack”), formed by normal-appearance phrases that we aim to avoid. In our pipeline, $y_n$ is encoded and used as unconditional embedding. The noise is computed via standard classifier-free guidance. 
At each sampling step $t$, the noise estimate is
\begin{equation}
\label{eq:negprompt-main}
\hat{\epsilon}_t
=\epsilon_\theta(x_t, y_n, t)
+ g\,\big(\epsilon_\theta(x_t, y, t)-\epsilon_\theta(x_t, y_n, t)\big),
\end{equation}
where $g$ is the classifier-free guidance scale. See Appendix for more details.

\subsection{Dual-Attention Enhancement}
\label{sec:DAE}

During generation, we observe that the generated anomalies guided by the optimized anomaly text embedding and self-attention grafting module may not fully occupy the target mask, which limits downstream improvements in anomaly localization. To address this issue, we introduce Dual-Attention Enhancement (DAE) at selected timesteps to reweight both cross- and self-attention in defect regions, achieving mask-consistent synthesis.

Self-attention maps encode pairwise similarity among spatial pixels. We therefore encourage queries in the target-anomaly branch to preferentially attend to anomalous features inside reference mask $M_{\mathrm{R}}$ in reference-anomaly diffusion branch. Concretely, within the foreground anomalous area we add an additive logit bias $\gamma$ and apply a temperature $\tau_{\mathrm{fg}}$ scaling to amplify the attraction of masked pixels to anomaly features in the target image. In particular, we modify Eq.~\eqref{eq:fg} into Eq.~\eqref{eq:corss-attention-up} below:
\begingroup
\setlength{\abovedisplayshortskip}{4pt}
\setlength{\belowdisplayshortskip}{4pt}
\setlength{\jot}{3pt}                   
\begin{equation}
\label{eq:corss-attention-up}
\begin{aligned}
\hat{\boldsymbol{A}}_{\mathrm{fg},t,l}
&= \boldsymbol{A}_{\mathrm{fg},t,l} + \log(\gamma)\,\overline{M}_{R},\\[-1pt]
\mathrm{\hat{\boldsymbol{Attn}}}_{\mathrm{fg}}
&= \mathrm{softmax}\!\Bigl(
    \frac{\hat{\boldsymbol{A}}_{\mathrm{fg},t,l}}{\tau_{\mathrm{fg}}}
    \!+\! \mathcal{MF}(M_R{=}0,-\infty)
  \Bigr)\,\boldsymbol{V}_{R,t,l}.
\end{aligned}
\end{equation}
\endgroup
In addition, recall that a cross-attention map represents a probability distribution over text tokens for each image patch, which determines the dominant token for that patch. For prompts of the template “\textit{A photo of a [cls] with a [anomaly\_type]}”, we aim to strengthen the influence of the anomaly token \textit{[anomaly\_type]} within the target mask on the final image. We observe that, after optimization, the attributes of the anomaly token in the prompt embedding remain localized at their original token position without semantic leakage. Therefore, we scale the attention assigned to \textit{[anomaly\_type]} by a factor $C=100$ in Eq.~\eqref{eq:cross-edit} (while keeping all other token attentions unchanged), which enhances its effect on synthesis within the masked region.
\begingroup
\begin{equation}
\label{eq:cross-edit}
\mathcal{F}\!\left(\boldsymbol{A}_{\mathrm{c},t}, M_T\right)_{i,j}
= \vcenter{\hbox{$
\begin{cases}
C\!\cdot\! M_T \!\cdot\! \big(\boldsymbol{A}_{\mathrm{c},t}\big)_{i,j}, & j = j^{*},\\[2pt]
\big(\boldsymbol{A}_{\mathrm{c},t}\big)_{i,j}, & \text{otherwise.}
\end{cases}
$}}
\end{equation}
\endgroup
Index $i$ represents the pixel value, while column index $j$ corresponds to text tokens, with $j^{*}$ indicating the position of the \textit{[anomaly\_type]} token. We gate DAE by timestamp $\tau_s$ for the self-attention enhancement and $\tau_c$ for the cross-attention enhancement.

Finally, the overall \method workflow is summarized in Algorithm~\ref{alg:triag} in Appendix~\ref{sec:Algorithmic}. In our setting, TriAG performs self-attention editing after the object shape has emerged, starting at denoising step \(T_S\), and at the designated layers \(\ell \in L_S\). DAE module comprises self- and cross-attention enhancements: cross-attention is triggered for \(t \in \tau_c\), whereas self-attention is triggered for \(t \in \tau_s\). Accordingly, at denoising step \(t\) and self-attention layer $\ell$, the attention \textbf{EDIT} function in Algorithm~\ref{alg:triag} is given by:
\begingroup
\begin{equation}
\label{eq:attention-edit}
\vcenter{\hbox{$
\mathrm{\textbf{EDIT}} := 
\left\{
\begin{array}{@{}c@{\quad}l@{}}
\mathrm{TriAG}, & \text{if } t > T_{S} \text{ and } \ell \in \mathrm{L}_{S},\\[2pt]
\mathrm{DAE},   & \text{if } t \in \tau_s \text{ or } t \in \tau_c,\\[2pt]
\multicolumn{2}{@{}l@{}}{\mathit{Self\mbox{-}Attention}(\{\boldsymbol{Q}_T, \boldsymbol{K}_T, \boldsymbol{V}_T\}),\ \text{otherwise}.}
\end{array}
\right.
$}}
\end{equation}
\endgroup
$\mathit{Self\mbox{-}Attention}$ stands for the standard self-attention operation~\cite{vaswani2017attention}. Importantly, without any fine-tuning, \method leverages rich knowledge inside the SD Model to perform same-category anomalous feature transfer to produce diverse, realistic anomalies. Moreover, it can scale to cross-category anomaly synthesis under zero-shot settings to enlarge the pool of reference anomalous data(see Sec.~\ref{sec:zero-shot}).


\section{Experiment}
\subsection{Experimental Settings}
\label{sec:settings}

\noindent\textbf{Datasets.}
We conduct experiments on the MVTec-AD dataset~\cite{bergmann2019mvtec}, which contains 15 product categories, each with up to eight anomaly types. Following the few-shot experimental setting of DualAnoDiff~\cite{jin2025dual} and related work, we use the first one-third of the images as reference anomaly images and evaluate on the remaining two-thirds.

\noindent\textbf{Evaluation Metrics.}
\textbf{1) Anomaly Generation.}We evaluate generative quality and diversity with Kernel Inception Distance (KID)~\cite{binkowski2018demystifying} and Intra-cluster pairwise LPIPS distance (IC-LPIPS)~\cite{ojha2021few}. KID for evaluating image quality is an unbiased estimator and more reliable than FID~\cite{heusel2017gans} on small datasets. \textbf{2) Anomaly Inspection.} We evaluate detection and localization with AUROC, AP and $F_1$-max score, and report Accuracy for anomaly classification.

\noindent\textbf{Implementation Details.}
We deploy the pre-trained Stable Diffusion v1.5~\cite{rombach2022high} with 50 denoising steps for anomaly synthesis without additional training. During \textsc{AGO}, we optimize the anomaly text embedding for 500 steps on the \(64\times64\) diffusion stage using Adam~\cite{kingma2017adammethodstochasticoptimization} with a fixed learning rate of \(3\times10^{-3}\). More details are supplied in Appendix.

\subsection{Comparison in Anomaly Generation}

\noindent\textbf{Baseline.}
We compare our method with existing anomaly-generation approaches. The most relevant training-free methods are TF\textsuperscript{2}~\cite{yu2024tf2} and AnomalyAny~\cite{sun2025unseen}. However, TF\textsuperscript{2} has no public code and reports only KID and classification accuracy and AnomalyAny lacks their designed text prompts and does not specify mask generation procedure for downstream anomaly detection.  Accordingly, we mainly compare against training-based few-shot anomaly generation methods—DFMGAN~\cite{duan2023few}, AnomalyDiffusion~\cite{hu2024anomalydiffusion}, DualAnoDiff~\cite{jin2025dual}, and SeaS~\cite{dai2025seasfewshotindustrialanomaly}—evaluated along three tasks: (i) anomaly-generation quality, (ii) anomaly detection and localization, and (iii) classification. For completeness, we present the detection results given by AnomalyAny in Tab.~\ref{tab:training_free} and include TF\textsuperscript{2} only in the classification comparison, using the results reported in their paper.

\begin{table*}[t]
\centering
\small
\caption{Comparison on a trained U-Net segmentation model~\cite{ronneberger2015u} for anomaly detection and localization on MVTec-AD dataset.}
\label{tab:mvtec_detection}
\vspace{-0.4em}
\setlength{\tabcolsep}{3.2pt}
\resizebox{\textwidth}{!}{
\begin{tabular}{c
|cccc
|cccc
|cccc
|cccc
|cccc}
\toprule
\multirow{2}{*}{Category} &
\multicolumn{4}{c|}{\textbf{DFMGAN~\cite{duan2023few}}} &
\multicolumn{4}{c|}{\textbf{AnomalyDiffusion~\cite{hu2024anomalydiffusion}}} &
\multicolumn{4}{c|}{\textbf{DualAnoDiff~\cite{jin2025dual}}} &
\multicolumn{4}{c|}{\textbf{SeaS~\cite{dai2025seasfewshotindustrialanomaly}}} &
\multicolumn{4}{c}{\textbf{Ours}} \\
& AP-I & AUC-P & AP-P & F1-P
& AP-I & AUC-P & AP-P & F1-P
& AP-I & AUC-P & AP-P & F1-P
& AP-I & AUC-P & AP-P & F1-P
& AP-I & AUC-P & AP-P & F1-P \\
\midrule
bottle
& 99.8 & 98.9 & 90.2 & 83.9
& \underline{99.9} & 99.4 & 94.1 & 87.3
& \textbf{100} & \underline{99.5} & 93.4 & 85.7
& \underline{99.9} & \textbf{99.7} & \textbf{95.9} & \textbf{88.8}
& \textbf{100} & \textbf{99.7} & \underline{95.4} & \underline{88.4} \\
cable
& 97.8 & 97.2 & 81.0 & 75.4
& \textbf{100} & \underline{99.2} & \underline{90.8} & \underline{83.5}
& 98.3 & 98.5 & 82.6 & 76.9
& \underline{98.8} & 96.0 & 83.1 & 77.7
& \textbf{100} & \textbf{99.4} & \textbf{91.2} & \textbf{85.0} \\
capsule
& 98.5 & 79.2 & 26.0 & 35.0
& \textbf{99.9} & \underline{98.8} & \underline{57.2} & \underline{59.8}
& 99.2 & \textbf{99.5} & \textbf{73.2} & \textbf{67.0}
& 99.2 & 93.7 & 41.9 & 47.3
& \underline{99.8} & 97.0 & \underline{60.6} & 59.0 \\
carpet
& 98.5 & 90.6 & 33.4 & 38.1
& 98.8 & 98.6 & 81.2 & 74.6
& \underline{99.9} & \underline{99.4} & \textbf{89.1} & \textbf{80.2}
& 99.0 & 99.3 & 86.4 & 78.1
& \textbf{100} & \textbf{99.5} & \underline{88.5} & \underline{80.0} \\
grid
& 90.4 & 75.2 & 14.3 & 20.5
& 99.5 & 98.3 & 52.9 & 54.6
& 99.7 & 98.5 & 57.2 & 54.9
& \underline{99.9} & \textbf{99.7} & \underline{76.3} & \underline{70.0}
& \textbf{100} & \underline{99.6} & \textbf{78.6} & \textbf{71.6} \\
hazelnut
& \textbf{100} & \underline{99.7} & 95.2 & 89.5
& \underline{99.9} & \textbf{99.8} & \underline{96.5} & \underline{90.6}
& \textbf{100} & \textbf{99.8} & \textbf{97.7} & \textbf{92.8}
& 99.8 & 99.5 & 92.3 & 85.6
& \textbf{100} & \textbf{99.8} & 96.2 & 90.1 \\
leather
& \textbf{100} & 98.5 & 68.7 & 66.7
& \textbf{100} & 99.8 & 79.6 & 71.0
& \textbf{100} & \textbf{99.9} & \textbf{88.8} & \underline{78.8}
& \textbf{100} & \underline{99.8} & 85.2 & 77.0
& \textbf{100} & 99.7 & \underline{88.0} & \textbf{79.7} \\
metal\_nut
& 99.8 & 99.3 & 98.1 & \underline{94.5}
& \textbf{100} & \textbf{99.8} & \underline{98.7} & 94.0
& \underline{99.9} & \underline{99.6} & 98.0 & 93.0
& \textbf{100} & \textbf{99.8} & \textbf{99.2} & \textbf{95.7}
& \textbf{100} & \textbf{99.8} & \textbf{99.2} & \textbf{95.7} \\
pill
& 91.7 & 81.2 & 67.8 & 72.6
& \underline{99.6} & \underline{99.7} & 93.9 & \textbf{90.8}
& 99.0 & 99.6 & 95.8 & 89.2
& \underline{99.6} & \textbf{99.9} & \textbf{97.1} & \underline{90.7}
& \textbf{99.7} & \underline{99.7} & \underline{96.1} & 89.9 \\
screw
& 64.7 & 58.8 & 2.2 & 5.3
& 97.9 & 97.0 & 51.8 & 50.9
& 95.0 & 98.1 & 57.1 & 56.1
& \underline{98.0} & \underline{98.5} & \underline{58.5} & \underline{57.2}
& \textbf{98.2} & \textbf{99.4} & \textbf{68.2} & \textbf{64.4} \\
tile
& \textbf{100} & 99.5 & 97.1 & 91.6
& \textbf{100} & 99.2 & 93.6 & 86.2
& \textbf{100} & 99.7 & 97.1 & 91.0
& \textbf{100} & \underline{99.8} & \underline{97.9} & \underline{92.5}
& \textbf{100} & \textbf{99.9} & \textbf{98.2} & \textbf{92.7} \\
toothbrush
& \textbf{100} & 96.4 & \underline{75.9} & \underline{72.6}
& \textbf{100} & \textbf{99.2} & \textbf{76.5} & \textbf{73.4}
& \textbf{99.7} & 98.2 & 68.3 & 68.6
& \textbf{100} & \underline{98.4} & 70.0 & 68.1
& \textbf{100} & 96.3 & 58.6 & 59.2 \\
transistor
& 92.5 & 96.2 & 81.2 & 77.0
& \textbf{100} & \underline{99.3} & \underline{92.6} & \underline{85.7}
& \underline{93.7} & 98.0 & 86.7 & 79.6
& 99.5 & 98.0 & 87.3 & 81.9
& \textbf{100} & \textbf{99.9} & \textbf{98.2} & \textbf{93.2} \\
wood
& 99.4 & 95.3 & 70.7 & 65.8
& 99.4 & \textbf{98.9} & 84.6 & 74.5
& \textbf{99.9} & 99.4 & \textbf{91.6} & \textbf{83.8}
& 99.6 & 99.0 & 87.0 & 79.6
& \underline{99.8} & \underline{99.4} & \underline{89.4} & \underline{81.1} \\
zipper
& \underline{99.9} & 92.9 & 65.6 & 64.9
& \textbf{100} & \underline{99.6} & 86.0 & 79.2
& \textbf{100} & \underline{99.6} & \textbf{90.7} & \textbf{82.7}
& \textbf{100} & \textbf{99.7} & 88.2 & \underline{81.6}
& \textbf{100} & 99.5 & \underline{88.5} & \underline{82.0} \\
\midrule[1.1pt]
\rowcolor{AvgRow}
\textbf{Average}
& 94.8 & 90.0 & 62.7 & 62.1
& 99.7 & 99.1 & 81.4 & 76.3
& 98.9 & 99.1 & 84.5 & 78.8
& 99.6 & 98.7 & 83.1 & 78.1
& \textbf{99.8} & \textbf{99.2} & \textbf{86.3} & \textbf{80.8} \\
\bottomrule[1.1pt]
\end{tabular}}
\end{table*}

\setlength{\textfloatsep}{6pt}
\setlength{\intextsep}{6pt}
\begin{table}[t]
\centering
\small
\setlength{\tabcolsep}{6pt}
\resizebox{\columnwidth}{!}{
\begin{tabular}{c|cc|cc|cc|cc|cc}
\toprule
\multirow{2}{*}{Category} &
\multicolumn{2}{c|}{DFMGAN~\cite{duan2023few}} &
\multicolumn{2}{c|}{AnoDiff~\cite{hu2024anomalydiffusion}} &
\multicolumn{2}{c|}{DualAno~\cite{jin2025dual}} &
\multicolumn{2}{c|}{SeaS~\cite{dai2025seasfewshotindustrialanomaly}} &
\multicolumn{2}{c}{Ours} \\
& KID$\downarrow$ & IC\!-L$\uparrow$
& KID$\downarrow$ & IC\!-L$\uparrow$
& KID$\downarrow$ & IC\!-L$\uparrow$
& KID$\downarrow$ & IC\!-L$\uparrow$
& KID$\downarrow$ & IC\!-L$\uparrow$ \\
\midrule
bottle      & \underline{70.90} & 0.12 & 137.71 & 0.17 & 151.09 & \textbf{0.31} & 148.29 & \underline{0.21} & \textbf{42.78} & 0.17 \\
cable       & \underline{61.68} & 0.25 &  88.44 & 0.40 &  86.01 & \textbf{0.45} & 108.62 & \textbf{0.43} & \textbf{28.93} & 0.41 \\
capsule     & \underline{40.63} & 0.11 & 111.30 & 0.18 &  72.82 & \underline{0.32} & 183.11 & 0.29 & \textbf{22.71} & 0.17 \\
carpet      & \textbf{25.14} & 0.13 & 107.87 & 0.21 &  56.80 & \textbf{0.27} & 127.62 & \underline{0.26} & \underline{45.10} & \textbf{0.29} \\
grid        & 101.04 & 0.13 & \underline{51.55} & \textbf{0.45} &  64.13 & \underline{0.44} &  60.85 & 0.41 & \textbf{7.04} & 0.38 \\
hazelnut    & \underline{21.16} & 0.24 &  65.70 & 0.28 &  40.20 & \textbf{0.37} &  46.96 & \underline{0.31} & \textbf{9.54} & 0.30 \\
leather     & \textbf{75.85} & 0.17 & 221.43 & \textbf{0.39} & 208.71 & \underline{0.35} & 199.18 & \textbf{0.39} & \underline{168.22} & 0.34 \\
metal\_nut  & \underline{44.04} & \underline{0.32} & 102.11 & 0.26 &  63.60 & \textbf{0.33} & 116.17 & \underline{0.32} & \textbf{43.07} & 0.29 \\
pill        & 123.52 & 0.16 & \underline{64.44} & 0.23 & 153.44 & \textbf{0.41} &  77.70 & \underline{0.31} & \textbf{39.76} & 0.24 \\
screw       & \underline{9.53} & 0.14 &  10.03 & 0.29 &  36.81 & \textbf{0.39} &  34.47 & \underline{0.33} & \textbf{1.48} & \underline{0.33} \\
tile        & \underline{85.28} & 0.22 & 181.28 & 0.48 & \textbf{57.95} & \underline{0.49} & 170.94 & \textbf{0.51} & 114.94 & 0.46 \\
toothbrush  & 46.49 & 0.18 & \underline{42.33} & 0.17 &  59.79 & \textbf{0.48} & 103.66 & \underline{0.31} & \textbf{10.75} & 0.20 \\
transistor  & \underline{88.31} & 0.25 & 118.70 & 0.30 & 119.12 & \underline{0.32} & 105.39 & \textbf{0.36} & \textbf{46.41} & \underline{0.32} \\
wood        & 68.13 & 0.35 &  74.62 & 0.35 & \underline{52.17} & \underline{0.39} & 130.96 & \textbf{0.48} & \textbf{36.64} & 0.38 \\
zipper      & \underline{78.08} & 0.27 & 162.58 & 0.24 & 335.47 & \textbf{0.41} & 271.11 & \underline{0.30} & \textbf{65.84} & 0.25 \\
\midrule[1.1pt]
\rowcolor{AvgRow}
\textbf{Average}
            & \underline{62.85} & 0.20 & 102.67 & 0.29 & 103.87 & \textbf{0.38} & 125.67 & \underline{0.35} & \textbf{45.55} & 0.30 \\
\bottomrule[1.1pt]
\end{tabular}}
\caption{Comparison on KID and IC-LPIPS on MVTec dataset. Bold and underline represent optimal and sub-optimal results, respectively. Our method generates anomaly images closest to the real distribution, achieving the lowest KID. Although our IC-LPIPS is lower than some methods, their higher IC-LPIPS partly results from corrupting the normal background.}
\label{tab:kid_icl}
\end{table}

\noindent\textbf{Anomaly image generation quality.}
Representative anomaly image–mask pairs are shown in Fig.~\ref{fig:generated_anomaly}, and the quantitative results are summarized in Tab.~\ref{tab:kid_icl}. We assess fidelity with Kernel Inception Distance (KID; lower is better) and measure diversity with IC-LPIPS (higher is better). In our setup, anomaly masks are generated from AnomalyDiffusion~\cite{hu2024anomalydiffusion}. As reported in Tab.~\ref{tab:kid_icl}, our method attains lower KID across categories, indicating a closer match to the real anomaly distribution. Our lower IC-LPIPS mainly results from synthesizing anomalies on simply augmented normal images from limited training set and superior preservation of normal background outside the anomalous region. By contrast, methods such as DualAnoDiff~\cite{jin2025dual} may bring background corruptions, inflating IC-LPIPS while compromising fidelity as illustrated in Fig.~\ref{fig:generated_anomaly}.

\noindent\textbf{Anomaly generation for anomaly detection and localization.}
We evaluate the effectiveness of our generated data for anomaly detection and localization, comparing against existing anomaly generation methods. For each method, we synthesize 1,000 image–mask pairs to train a U-Net segmentation model~\cite{ronneberger2015u}, following the protocol established in AnomalyDiffusion~\cite{hu2024anomalydiffusion} and DualAnoDiff~\cite{jin2025dual}. We report image-level AP (AP-I) and pixel-level metrics (AUROC-P, AP-P and F1-P in Tab.~\ref{tab:mvtec_detection}, with additional image-level results (AUROC-I, F1-max) in the Appendix~\ref{app-G}. Our method attains the best performance across all reported metrics.

\noindent\textbf{Anomaly generation for anomaly classification.}
To measure the quality of our synthesized data for downstream tasks, we train a ResNet-34~\cite{he2016deep} classifier on generated images under AnomalyDiffusion~\cite{hu2024anomalydiffusion} protocol. As reported in Tab.~\ref{tab:cls_acc}, our method achieves the best classification accuracy across categories, surpassing the previous state-of-the-art (Anomalydiffusion~\cite{hu2024anomalydiffusion}) by 12.1\% points, indicating our synthesized anomaly images are more realistic. Moreover, our \method outperforms the training-free TF\textsuperscript{2}~\cite{yu2024tf2} by 20.47\% points. We attribute this gap to TF\textsuperscript{2} transferring defect features at the latent feature level (coarser and less semantically precise), whereas our approach performs fine-grained attention editing, yielding more faithful anomalies aligning with real distribution.

{%
\setlength{\abovecaptionskip}{8.2pt}  
\begin{figure}[t] 
  \centering
  \includegraphics[width=\columnwidth]{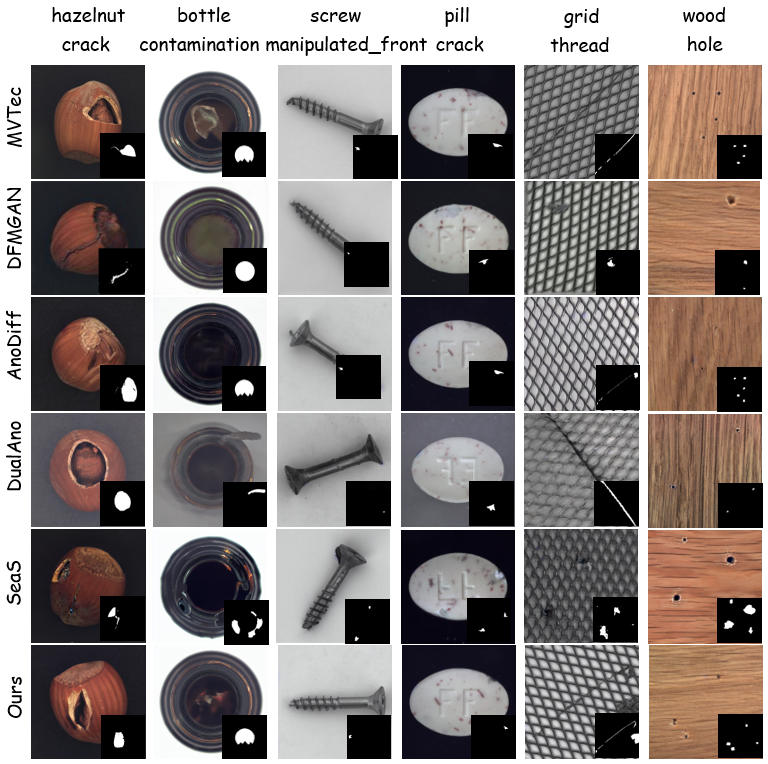} 
  \caption{Qualitative comparison of generated results on MVTec-AD. The sub-image in the lower right corner is anomaly mask.}
  \label{fig:generated_anomaly}
\end{figure}
}

\subsection{Zero-shot cross-class Generation}
\label{sec:zero-shot}

As discussed in \autoref{sec:TriAG}, self-attention encodes rich defect information, and the anomaly attributes tied to the same \textit{[anomaly\_type]} can transfer across object categories \textit{[cls]}. We evaluate the zero-shot ability of \method (Fig.~\ref{fig:cross}) by using all \emph{wood-hole} test images as reference images to guide the synthesis of \emph{hazelnut-hole}. During generation we only access anomaly images from \emph{wood-hole} and normal images from \emph{hazelnut}.

We also test whether baselines can operate in a zero-shot setting. Among them, only SeaS~\cite{dai2025seasfewshotindustrialanomaly} and AnomalyAny~\cite{sun2025unseen} supports this setting. SeaS is designed to use unbalanced abnormal prompt 
\(P=\) ``a \texttt{<ob>} with \texttt{<df$_1$>}, \texttt{<df$_2$>}, \ldots, \texttt{<df$_N$>}'' (\(N{=}4\)), where \texttt{<ob>} captures the normal \textit{[cls]} appearance and \texttt{<df$_n$>} encodes the \textit{[anomaly\_type]} features. For cross-category transfer, we retrain SeaS with \emph{wood--hole} anomaly images and \emph{hazelnut} normal images so that the anomaly token \texttt{<df$_n$>} binds the \emph{hole} defect and the normal token \texttt{<ob>} captures category appearance. We further compare with the copy-paste method DRAEM~\cite{zavrtanik2021draem}. As shown in Fig.~\ref{fig:cross}, our \method transfers the \emph{hole} defect from \emph{wood} to \emph{hazelnut} while preserving the normal appearance outside the mask. We also train a U-Net on the synthesized image–mask pairs for detection. As shown in Table~\ref{tab:zero_shot}, although this zero-shot setting falls short of training with real anomaly samples, it still delivers competitive performance.

{
\setlength{\floatsep}{1pt}                  


\begin{table}[t]
\centering
\footnotesize                       
\setlength{\tabcolsep}{5pt}         
\renewcommand{\arraystretch}{0.98}  
\setlength{\aboverulesep}{0.2pt}      
\setlength{\belowrulesep}{0.2pt}
\caption{Comparison to training\!-free AnomalyAny on MVTec\!-AD.}
\label{tab:training_free}
\vspace{-0.42em}
\begin{tabular}{@{}lcccc@{}}
\toprule
Metric & AUC-I & F1-I & AUC-P & F1-P \\
\midrule
AnomalyAny       & 98.4 & 96.9 & 97.4 & 65.1 \\
Ours             & 99.6 (+\textbf{1.2}) & 99.2 (+\textbf{2.3}) & 99.2 (+\textbf{1.8}) & 80.6 (+\textbf{15.5}) \\
\bottomrule
\end{tabular}
\end{table}
\begin{table}[t]
\setlength{\textfloatsep}{4pt}
\centering
\small
\setlength{\tabcolsep}{6pt}
\caption{Anomaly classification accuracy (\%) with ResNet-34~\cite{he2016deep} trained on images synthesized by each method.}
\label{tab:cls_acc}
\vspace{-0.42em}
\resizebox{\columnwidth}{!}{
\begin{tabular}{c|c|c|c|c|c|c}
\toprule
Category & DFMGAN~\cite{duan2023few} & AnoDiff~\cite{hu2024anomalydiffusion} & TF\textsuperscript{2}~\cite{yu2024tf2} & DualAno~\cite{jin2025dual} & SeaS~\cite{dai2025seasfewshotindustrialanomaly} & Ours \\
\midrule
bottle     & 56.59 & \textbf{88.37} & 79.36 & 67.44 & 81.40 & \underline{86.05} \\
cable      & 45.31 & \textbf{76.56} & \underline{60.00} & 57.81 & 48.44 & \textbf{76.56} \\
capsule    & 37.23 & 44.00 & \underline{53.12} & 50.67 & 33.33 & \textbf{77.33} \\
carpet     & 47.31 & 58.06 & 55.05 & \underline{62.90} & 38.71 & \textbf{74.19} \\
grid       & 40.83 & 60.00 & 47.36 & \underline{67.50} & 47.50 & \textbf{80.00} \\
hazelnut   & 81.94 & 81.25 & \underline{88.88} & 79.17 & 81.25 & \textbf{95.83} \\
leather    & 49.73 & 65.08 & 46.06 & \underline{84.13} & 61.90 & \textbf{92.06} \\
metal\_nut & 64.58 & \underline{82.81} & 72.04 & 73.44 & 68.75 & \textbf{84.38} \\
pill       & 29.52 & \underline{64.58} & 34.75 & 41.67 & 18.75 & \textbf{72.92} \\
screw      & 37.45 & 29.63 & \underline{40.33} & 39.51 & \textbf{56.79} & \textbf{56.79} \\
tile       & 74.85 & \underline{92.98} & 88.09 & \textbf{100.0} & 89.47 & \textbf{100.0} \\
transistor & 52.38 & 75.00 & 67.50 & \underline{78.57} & 57.14 & \textbf{92.86} \\
wood       & 49.21 & \underline{78.57} & 70.00 & \textbf{90.48} & 76.19 & \underline{78.57} \\
zipper     & 27.64 & \textbf{86.59} & 63.86 & 24.39 & 34.15 & \underline{85.37} \\
\midrule[1.1pt]
\rowcolor{AvgRow}
\textbf{Average} & 49.61 & \underline{70.25} & 61.88 & 65.55 & 56.70 & \textbf{82.35} \\
\bottomrule[1.1pt]
\end{tabular}}
\end{table}
}

{
\setlength{\abovecaptionskip}{8pt}  
\begin{figure}[t] 
  \centering
  \includegraphics[width=\columnwidth]{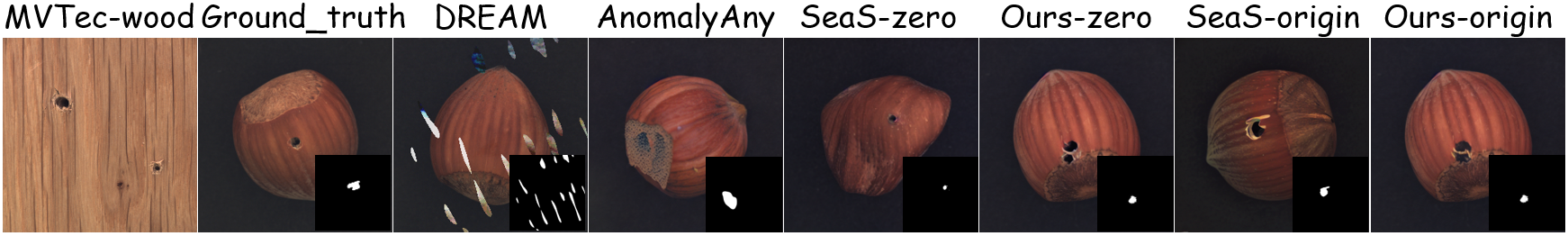} 
  \caption{Qualitative comparison of generated \emph{hazelnut-hole}.}
  \label{fig:cross}
\end{figure}
}

\subsection{Ablation Study}
We evaluate the effectiveness of the three components: Tri-branch Attention Grafting (TriAG), Dual-Attention Enhancement (DAE), and Anomaly-Guided Optimization (AGO). Specifically, we consider four configurations: 1) TriAG only; 2) TriAG + DAE; 3) TriAG + AGO; and 4) the full framework. For each configuration and anomaly type, we synthesize 1000 images and train a U-Net segmentation model for pixel-level detection. Table~\ref{tab:ablation} shows that removing any proposed module degrades detection performance, confirming the effectiveness of each component. Additional experimental details are provided in the Appendix.

{
\setlength{\floatsep}{1pt}               
\begin{table}[t]
\centering
\small
\setlength{\belowcaptionskip}{4pt}
\renewcommand{\arraystretch}{0.91}  
\setlength{\tabcolsep}{6pt}
\caption{Comparison between zero-shot and original settings on \emph{hazelnut–hole} subset of MVTec-AD for anomaly detection.}
\label{tab:zero_shot}
\vspace{-0.42em}
\resizebox{\columnwidth}{!}{
\begin{tabular}{l|ccc|ccc}
\toprule
\multirow{2}{*}{Method} & \multicolumn{3}{c|}{Image-level} & \multicolumn{3}{c}{Pixel-level} \\
& AUC-I & AP-I & F1-I & AUC-P & AP-P & F1-P \\
\midrule
DREAM                & 99.6 & 99.1 & 94.7 & 99.3 & 74.8 & 73.3 \\
AnomalyAny           & 100.0 & 100.0 & 100.0 & 98.6 & 78.8 & 75.0 \\
SeaS-zero           & 100.0 & 100.0 & 100.0 & 97.5 & 78.6 & 73.7 \\
Ours-zero           & 100.0 & 100.0 & 100.0 & 99.1 & 86.8 & 80.2 \\
SeaS-origin         & 100.0 & 100.0 & 100.0 & 99.2 & 87.3 & 80.0 \\
Ours-origin         & 100.0 & 100.0 & 100.0 & 99.1 & 90.4 & 84.8 \\
\bottomrule
\end{tabular}}
\end{table}
\begin{table}[t]
\centering
\footnotesize
\setlength{\tabcolsep}{4.5pt}        
\renewcommand{\arraystretch}{0.90}   
\caption{Ablation on components. \cmark\ denotes enabled.}
\label{tab:ablation}
\vspace{-0.42em}
\begin{tabular*}{0.98\columnwidth}{@{\extracolsep{\fill}} ccc | ccc @{}}
\toprule
\multicolumn{3}{c|}{\textbf{Method}} & \multicolumn{3}{c}{\textbf{Metric}} \\
\textbf{TriAG} & \textbf{DAE} & \textbf{AGO} &
\textbf{AUROC}$\uparrow$ & \textbf{AP}$\uparrow$ & \textbf{F1-max}$\uparrow$ \\
\midrule
\cmark &        &        & 99.0   & 81.9   & 77.6   \\
\cmark & \cmark &        & 99.0 & 83.3 & 77.9 \\
\cmark &        & \cmark & 99.1 & 82.7 & 77.7 \\
\cmark & \cmark & \cmark & 99.2 & 86.3 & 80.8 \\
\bottomrule
\end{tabular*}
\end{table}
}

\section{Conclusion}

We propose \method, a training-free few-shot anomaly generation method. Motivated by the observation that anomalous foreground and normal background are separable in self-attention, \method coordinates attention across three parallel diffusion branches and applies mask-guided decoupling defects from background, producing text-consistent, realistic anomalies. To mitigate semantic drift, we introduce a lightweight Anomaly-Guided Optimization that reinforcing anomaly semantics. Our training-free method also allows using anomalies of the same anomaly type from other object categories to guide synthesis. Without model training, \method synthesizes distribution-faithful anomalies and achieves superior downstream detection performance over state-of-the-art training-based methods. 

\section*{Acknowledgments}

This work was supported by the National Natural Science Foundation of China (No. 62476124), the Fundamental and Interdisciplinary Disciplines Breakthrough Plan of the Ministry of Education of China (JYB2025XDXM902), the Natural Science Foundation of Jiangsu Province (No. BK20242015), the Gusu Innovation and Entrepreneur Leading Talents (No. ZXL2025322), the Fundamental Research Funds for the Central Universities (KG2025XX), the Jiangsu Science and Technology Major Project (BG2025035) and Nanjing University - China Mobile Communications Group Co., Ltd. Joint Institute. This research was also supported by cash and in-kind funding from the Nanjing Kunpeng\&Ascend Center of Cultivation and industry partner(s).
{
    \small
    \bibliographystyle{ieeenat_fullname}
    \bibliography{main}
}
\clearpage
\appendix
\setcounter{page}{1}
\maketitlesupplementary

\begin{algorithm*}[t]
\caption{Pipeline of Anomaly Generation in \method}
\label{alg:triag}
\begin{algorithmic}[1]
\Require A reference image--mask pair $(I_{\mathrm{R}}, M_{\mathrm{R}})$, a normal image $I_{\mathrm{N}}$, a target anomaly mask $M_{\mathrm{T}}$, and an anomaly text prompt $y$.
\Ensure Generated anomaly image $\hat{I}$.
\State $e^\star \gets \mathrm{TextEncoder}(y) - \nabla_{e}\,L_{\text{recon}}\!\big(I_{\mathrm{R}},\,\hat{I}_{\mathrm{R}}\big)$ \Comment{AGO}
\State $\{\boldsymbol{Z}_T^{\mathrm{ref}}, \boldsymbol{Z}_{T-1}^{\mathrm{ref}}, \ldots, \boldsymbol{Z}_0^{\mathrm{ref}}\} \gets \mathrm{Inversion}(I_{\mathrm{R}})$
\State $\{\boldsymbol{Z}_T^{\mathrm{nor}}, \boldsymbol{Z}_{T-1}^{\mathrm{nor}}, \ldots, \boldsymbol{Z}_0^{\mathrm{nor}}\} \gets \mathrm{Inversion}(I_{\mathrm{N}})$
\State $\boldsymbol{Z}_T^{\mathrm{tar}} \gets \boldsymbol{Z}_T^{\mathrm{nor}}$
\For{$t = T, T\!-\!1, \ldots, 1$}
  \State $\{\boldsymbol{Q}_R, \boldsymbol{K}_R, \boldsymbol{V}_R\} \gets \epsilon_{\theta}(\boldsymbol{Z}_t^{\mathrm{ref}}, t)$ \Comment{reference-anomaly diffusion branch}
  \State $\{\boldsymbol{Q}_N, \boldsymbol{K}_N, \boldsymbol{V}_N\} \gets \epsilon_{\theta}(\boldsymbol{Z}_t^{\mathrm{nor}}, t)$ \Comment{normal-image diffusion branch}
  \State $\{\boldsymbol{Q}_T, \boldsymbol{K}_T, \boldsymbol{V}_T\} \gets \epsilon_{\theta}(\boldsymbol{Z}_t^{\mathrm{tar}}, t)$ \Comment{target-anomaly diffusion branch}
  \State $\boldsymbol{\mathrm{\boldsymbol{Attn}}}^{\star} \gets
  \mathrm{\textbf{EDIT}}(\{\boldsymbol{Q}_T,\boldsymbol{K}_R,\boldsymbol{V}_R\},
  \{\boldsymbol{Q}_T,\boldsymbol{K}_N,\boldsymbol{V}_N\},
  M_{\mathrm{R}}, M_{\mathrm{T}})$
  \Comment{edit attention as Eq.\eqref{eq:cross-edit} (DAE and TriAG)}
  \State $\varepsilon \gets \epsilon_{\theta}(\boldsymbol{Z}_t^{\mathrm{tar}}, t, \boldsymbol{\mathrm{Attn}}^{\star}, e^\star)$ \Comment{noise prediction}
  \State $\boldsymbol{Z}_{t-1}^{\mathrm{tar}} \gets \mathrm{SampleDDIM}(\boldsymbol{Z}_t^{\mathrm{tar}}, \varepsilon, t)$ \Comment{DDIM step}
\EndFor
\State $\hat{I} \gets \mathrm{Decode}(\boldsymbol{Z}_0^{\mathrm{tar}})$
\State \Return $\hat{I}$
\end{algorithmic}
\end{algorithm*}

\section*{Overview}
This supplementary material provides additional details and results complementing the main paper. 
Specifically:
\begin{itemize}
    \item Sec.~\ref{sec:Algorithmic} describes the algorithmic pipeline of \method for anomaly generation.
    \item Sec.~\ref{sec:Analysis_self} presents additional self-attention feature analyses, providing further interpretability for our TriAG moudle of self-attention editing design.
    \item Sec.~\ref{app-C} supplies more experimental details and computation consumption.
    \item Sec.~\ref{app-D} provides additional experimental results on zero-shot cross-class anomaly generation.
    \item Sec.~\ref{app-E} presents qualitative visual results for the ablation of each component.
    \item Sec.~\ref{app-F} demonstrates the performance of our \method on the VisA dataset under the same settings.
    \item Sec.~\ref{app-I} demonstrates the performance of our \method on the Real-IAD dataset under one-reference settings.
    \item Sec.~\ref{app-G} reports more qualitative and quantitative results of anomaly image generation on MVTec-AD dataset.
    \item Sec.~\ref{app-H} analyzes the limitations of our \method and discusses potential directions for future improvements.
\end{itemize}

\section{Algorithmic Description of \method}
\label{sec:Algorithmic}

Algorithm~\ref{alg:triag} outlines the \method pipeline for synthesizing anomalous samples. Given a reference anomaly image–mask pair \((I_{\mathrm{R}}, M_{\mathrm{R}})\), a normal image \(I_{\mathrm{N}}\), and a target anomaly mask \(M_{\mathrm{T}}\), our goal is to generate an anomaly image \(\hat{I}\) that preserves the normal appearance outside \(M_{\mathrm{T}}\) while producing realistic, text-consistent defects within \(M_{\mathrm{T}}\). We first align text conditioning with anomaly semantics via the Anomaly-Guided Optimization module (\autoref{sec:AGO}), which refines the input prompt to yield an anomaly-aligned text embedding. The pipeline then runs three parallel diffusion branches: a \emph{reference–anomaly} branch and a \emph{normal–background} branch that process noised versions of \(I_{\mathrm{R}}\) and \(I_{\mathrm{N}}\), respectively, and a \emph{target–anomaly} branch initialized with the noised latent of \(I_{\mathrm{N}}\) to better retain structure and background. During sampling, we manipulate attention across the three branches at selected timesteps and attention layers using the \textbf{EDIT} function (Eq.~\eqref{eq:cross-edit}). Our attention editing mechanism comprises (i) \emph{Tri-branch Attention Grafting} (TriAG) within self-attention, which injects anomalous cues from the reference branch and background cues from the normal branch into the target branch, and (ii) \emph{Dual-Attention Enhancement} (DAE), which amplifies both cross-attention and self-attention responses. Together, these operations yield anomalies faithful to the target distribution and improve downstream detection performance.

\section{Analysis on Self-attention for Anomaly Generation}
\label{sec:Analysis_self}
As shown in \autoref{sec:TriAG}, we apply principal component analysis (PCA) to the self-attention maps of anomalous images and visualize the top three principal components, where regions with similar structure are rendered in similar colors. We further present the query (Q), key (K), and the resulting self-attention maps $\boldsymbol{A}=\mathrm{softmax}\!\left(\frac{\boldsymbol{Q}\boldsymbol{K}^{\mathrm T}}{\sqrt{d_k}}\right)$ to elucidate TriAG’s self-attention grafting mechanism in \cref{figa2:self-attention-vis}. 

As shown in Fig.~\ref{figa2:self-attention-vis}, early layers exhibit attention aligned with the image’s semantic layout, grouping regions by object parts and capturing coarse structure. Deeper layers progressively shift toward higher-frequency content, revealing fine-grained, patch-level texture cues. The spatial resolution follows the U-Net path—downsampling from \(64\times64\) to \(8\times8\), then upsampling back to \(64\times64\). Because the down and middle block provides limited informative self-attention in our setting, we omit it and perform self-attention grafting primarily on the up block (layers 10–16).

{%
\begin{figure}[t]
  \centering
  \includegraphics[width=\linewidth]{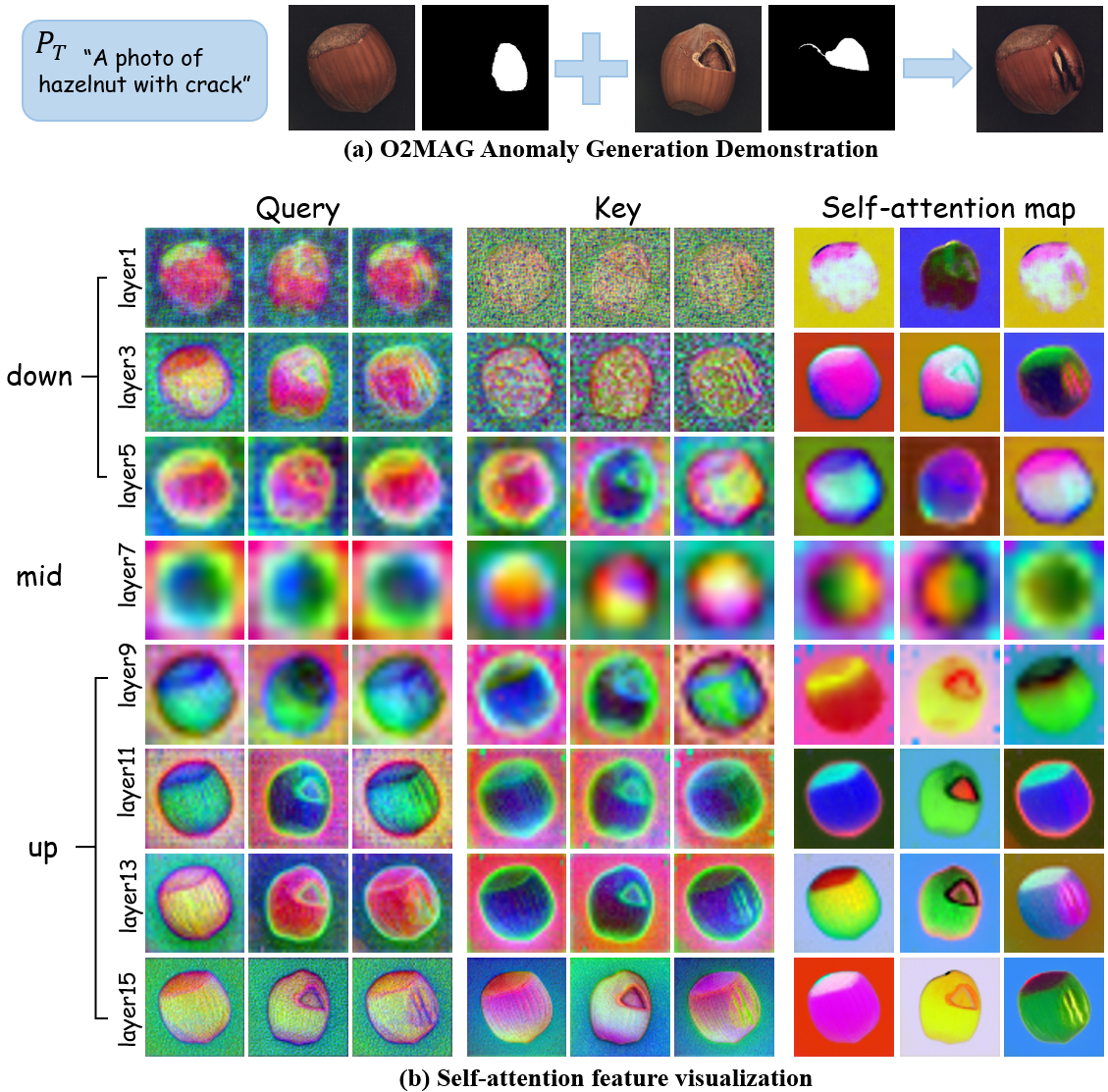} 
  \caption{(a) The Demonstration of \method Anomaly Generation, and (b) Self-attention feature maps visualized at the 30th denoising step. In each triplet of subfigures, the maps correspond (left$\rightarrow$right) to the normal image, the reference anomaly image, and the generated image.}
  \label{figa2:self-attention-vis}
\end{figure}
}

\begin{figure}[t]
  \centering
  \includegraphics[width=\linewidth]{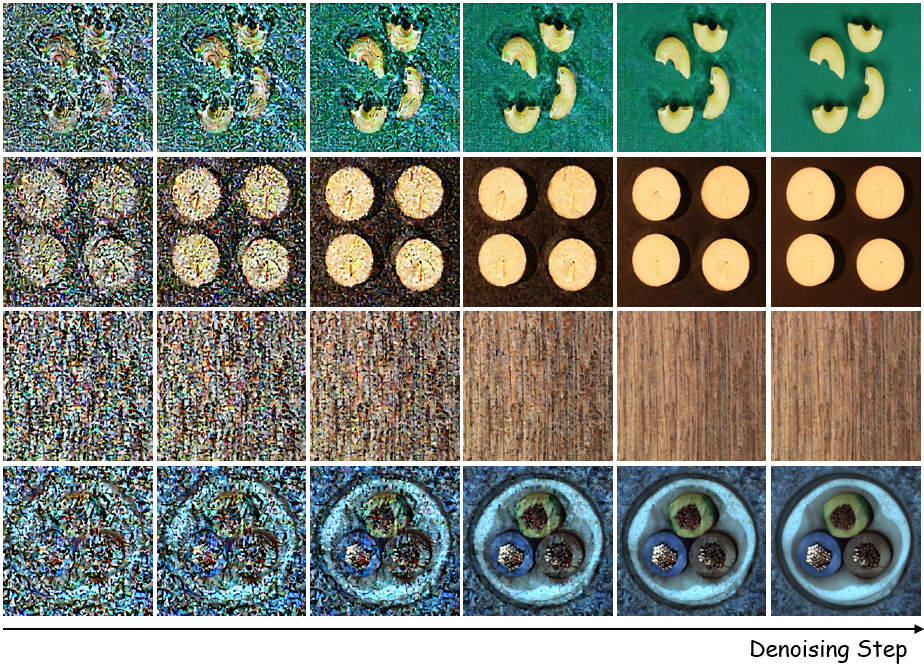} 
  \caption{The intermediate reconstruction during the iterative denoising process. For both texture and object categories, the global image layout is already formed in the early denoising steps.}
  \label{figa2:app-20}
\end{figure}

\section{More Experimental Details}
\label{app-C}

\subsection{More implementation details}
\noindent\textbf{Data preparation.}
We augment the normal images in the MVTec-AD training set (approximately 200–400 per class; toothbrush has only 60) using translations, flips, and rotations. For each category, we expand the training dataset of normal images to 1000 via data augmentation. For orientation-sensitive categories (e.g., zipper, capsule),  we use flips only to preserve canonical semantics. This lightweight augmentation, together with our method’s excellent preservation of normal background regions, may partly account for the lower IC-L compared with some baselines. Even so, our method still produces realistic and diverse anomalies.
We adopt anomaly masks generated by AnomalyDiffusion as the target masks. Because AnomalyDiffusion occasionally yields empty masks (all-zero), we generate 1{,}200 masks and retain 1{,}000 valid ones after filtering out the empty cases.

\noindent\textbf{Experimental Details and Hyperparameters.}
We deploy the pre-trained Stable Diffusion v1.5~\cite{rombach2022high} with 50 denoising steps and classifier-free guidance~\cite{ho2022classifier} set to 7.5 for anomaly synthesis without additional training. 

(1) \textbf{Additional Analysis on TriAG:} As illustrated in Fig.~\ref{figa2:app-20}, images in anomaly datasets exhibit relatively simple structure and their layout emerge early during denoising. Accordingly, we enable self-attention editing from the \(5th\) denoising step. Consistent with \autoref{sec:TriAG} and \autoref{sec:Analysis_self}, we perform attention grafting on the U-Net layers 10-16, which carry fine-grained, patch-level texture information.

\begin{figure}[h]
  \centering
  \includegraphics[width=0.98\linewidth]{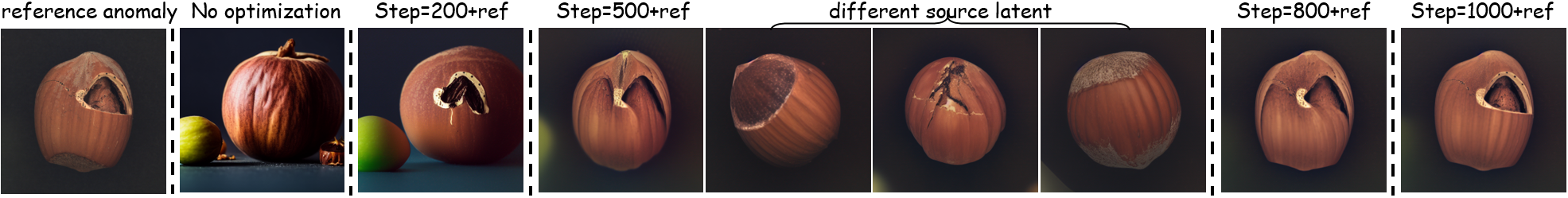}
  \vspace{-0.1cm}
  \caption{Analysis of the optimization step in AGO.}
  \label{fig:ago_steps}
\end{figure}

(2) \textbf{Additional Analysis on AGO:} During \textsc{AGO}, we optimize the anomaly text embedding for 500 steps on the \(64\times64\) diffusion stage using Adam with a fixed learning rate of \(3\times10^{-3}\). As illustrated in Fig.~\ref{fig:ago} of the main paper, optimization guides embeddings toward the target distribution. We set hyperparameters based on the reconstruction quality of the optimized prompt embeddings while accounting for computational cost, thereby balancing generation fidelity and optimization overhead. As illustrated in Fig.~\ref{fig:ago_steps}, our 500-step setting avoids the under-optimization of low steps while maintaining a substantial safety margin against the rigid reconstruction observed at 1000 steps.

Additionally, we explore incorporating data augmentation during the optimization stage. Specifically, we apply random rotation and translation to the reference anomaly images, augmenting each reference sample into five variants. We then optimize the corresponding anomaly text embeddings. Subsequently, during the generation process conditioned on the real reference images, we randomly sample an augmented image along with its optimized embedding to guide the generation. Due to the inherent similarity in object appearance and defect characteristics within anomaly datasets, the model still primarily focuses on learning normal background features and foreground anomalies even after augmentation. Consequently, the underlying appearance remains essentially unchanged, meaning that the overall diversity is not significantly increased. Instead, the primary role of this augmentation module is to enhance the realism of the generated samples, ensuring they better align with the true distribution of real-world anomaly data. The comparison results on hazelnut of MVTec-AD are presented in Tab.~\ref{tab:origin_augment}.

\begin{table}[t]
\centering
\setlength{\tabcolsep}{5pt}
\renewcommand{\arraystretch}{1.05}
\caption{Comparison between the original and augmented settings.}
\label{tab:origin_augment}
\resizebox{\columnwidth}{!}{
\begin{tabular}{l|ccc|ccc}
\toprule
Setting & AUC-I & AP-I & F1-I & AUC-P & AP-P & F1-P \\
\midrule
Origin  & 99.9 & 100.0 & 99.0 & 99.8 & 96.2 & 90.1 \\
Augment & 99.9 & 99.9  & 98.9 & 99.8 & 95.9 & 91.0 \\
\bottomrule
\end{tabular}
}
\end{table}

\begin{figure}[h]
  \centering
  \includegraphics[width=0.98\linewidth]{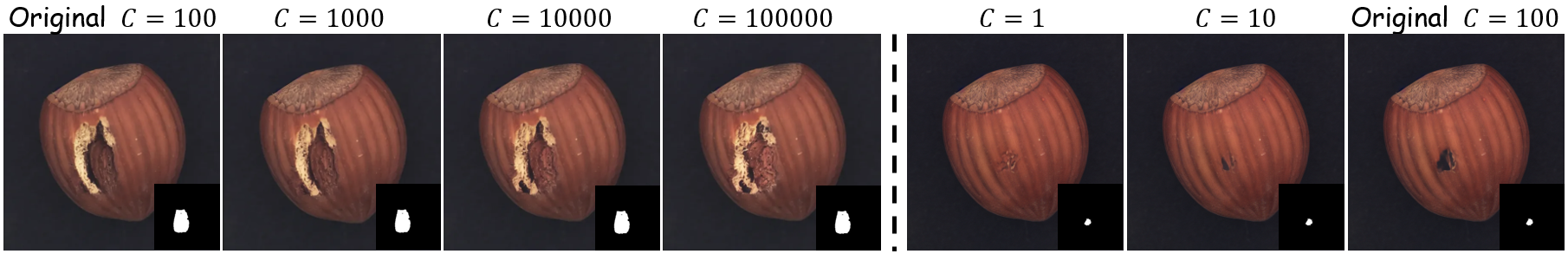}
  \vspace{-0.1cm}
  \caption{Analysis of the scaling factor $C$ in DAE.}
  \label{fig:dae_c}
\end{figure}

(3) \textbf{Additional Analysis on DAE:} The self-attention bias $\gamma$ is set to \(1.1\), the temperature $\tau_{\mathrm{fg}}$ to \(0.7\), and the cross-attention upweight to \(C=100\). Timestamp \(\tau_s\in(5,50)\) for self\hyp{}attention enhancement and \(\tau_c\in(20,40)\) for cross\hyp{}attention enhancement. We selected DAE hyperparameters via empirical validation to handle the trade-off between sensitivity and stability. While a sufficiently large amplification factor $C$ is essential for manifesting anomalies in small masks, Fig.~\ref{fig:dae_c} reveals a wide safety margin, with artifacts only emerging when $C > 10,000$. Similarly, $\gamma$ and $\tau_{fg}$ perform robustly across broad ranges.

Critically, we employ the \textbf{same} hyperparameters across all datasets (MVTec-AD, VisA, Real-IAD) without per-dataset tuning, validating the \textbf{robust generalizability} of our method.

In addition, we use a unified simple text prompt template, “\textit{A photo of a [cls] with a [anomaly\_type]}”, to specify the object class and anomaly type, which serves as the positive prompt. We further introduce negative prompts during synthesis. Technically, the positive prompt steers the diffusion process toward images consistent with its description, while the negative prompt pushes it away from the specified attributes. For anomaly types that are semantically related in the dataset, we construct antonym-like phrases as negative prompts. For example, for anomalies such as \textit{crack}, \textit{scratch}, and \textit{rough} in MVTec-AD, we adopt phrases describing normal appearance (e.g., “no crack”, “no cut”, “no scratch”, “smooth surface”) as negative prompts, encouraging the model to generate anomalies that deviate from the normal appearance.

\subsection{Resource requirement and computation consumption}
\label{sec:Resource_requirement}
We conduct anomaly synthesis on NVIDIA RTX 5880 Ada (48\,GB) GPUs for each product category, which may use about 30G memory. Table~\ref{tab:time_speed} reports runtime for training-based methods (DFMGAN, AnomalyDiffusion, DualAnoDiff, SeaS) and training-free methods (AnomalyAny and our \method). DFMGAN, DualAnoDiff, and SeaS need to train a separate model per category, which increases both training time and storage. In addition, DFMGAN and AnomalyDiffusion operates at \(256\times256\) resolution, whereas the other methods synthesize \(512\times512\) images. Within the training-free setting, AnomalyAny requires about 120\,s per image at inference, while our \method needs only 28\,s per image, achieving a \(\approx 4.3\times\) speedup.



\begin{table}[t]
\centering
\small
\setlength{\tabcolsep}{8pt}
\caption{Comparison of training cost and inference speed.}
\label{tab:time_speed}
\resizebox{\columnwidth}{!}{
\begin{tabular}{l|c|c} 
\toprule
\textbf{Methods} & \textbf{Training Overall Time} & \textbf{Inference Time (per image)} \\
\midrule
DFMGAN~\cite{duan2023few}                     & 464 hours       & 0.9\,s \\
AnomalyDiffusion~\cite{hu2024anomalydiffusion}& 310 hours       & 7.4\,s \\
DualAnoDiff~\cite{jin2025dual}                & 197 hours               & 32.4\,s \\
SeaS~\cite{dai2025seasfewshotindustrialanomaly}& 124\,hours & 10.7\,s \\
\midrule 
AnomalyAny~\cite{yu2024tf2}                   & 0 hours        & 120s \\
\method\ (ours)                               & 0 hours        & 28s \\
\bottomrule
\end{tabular}}
\end{table}


\section{Additional Zero-Shot Generation Details}
\label{app-D}

\begin{figure*}[t]
  \centering
  \includegraphics[width=\textwidth]{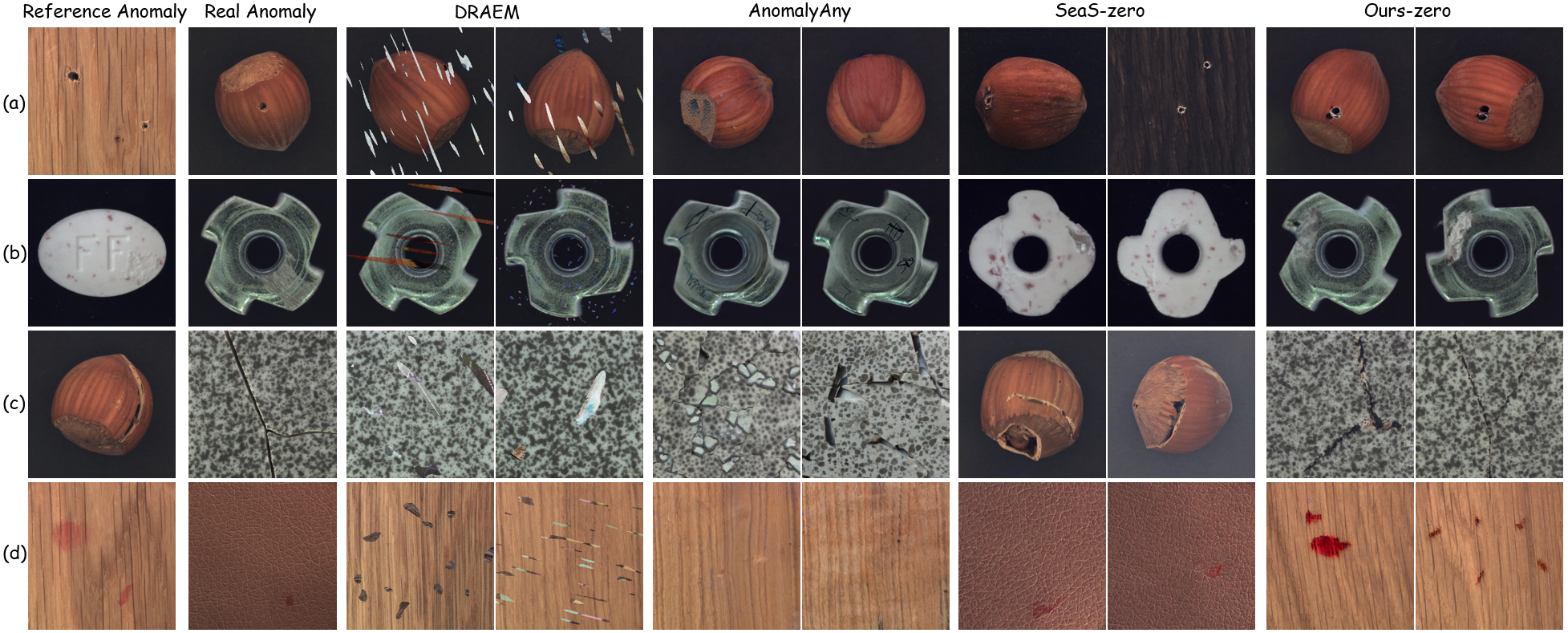}
  \caption{Anomaly generation under zero-shot settings. We aim to transfer anomalous features of the same anomaly type from “Reference Anomaly” images of other categories to the target category, synthesizing realistic defects that are consistent with "Real Anomaly".}
  \label{fig:app-21}
\end{figure*}

As discussed in \autoref{sec:Attention_Mechanism}, we evaluate the zero-shot generation capability of SeaS, AnomalyAny, and \method. In this setting, the generation model only accesses normal images from the target category and anomalous images from other categories without access to target category anomalies. In Fig.~\ref{fig:cross}, we transfer \textit{wood–hole} anomalous attributes to synthesize \textit{hazelnut–hole}. We further test other cross-class pairs including \textit{hazelnut–crack}\(\rightarrow\)\textit{tile–crack}, \textit{pill–scratch}\(\rightarrow\)\textit{metal\_nut–scratch}, and \textit{leather–color}\(\rightarrow\)\textit{wood–color}. Additional visualizations are shown in Fig.~\ref{fig:app-21}. Because self-attention features encode both texture and appearance information, attention-based feature transfer can cause appearance leakage into the target images, yielding cut-and-paste-like pseudo anomalies rather than seamlessly integrated defects.

\section{More Ablation Studies}
\label{app-E}

\begin{figure*}[t]
  \centering
  \includegraphics[width=\textwidth]{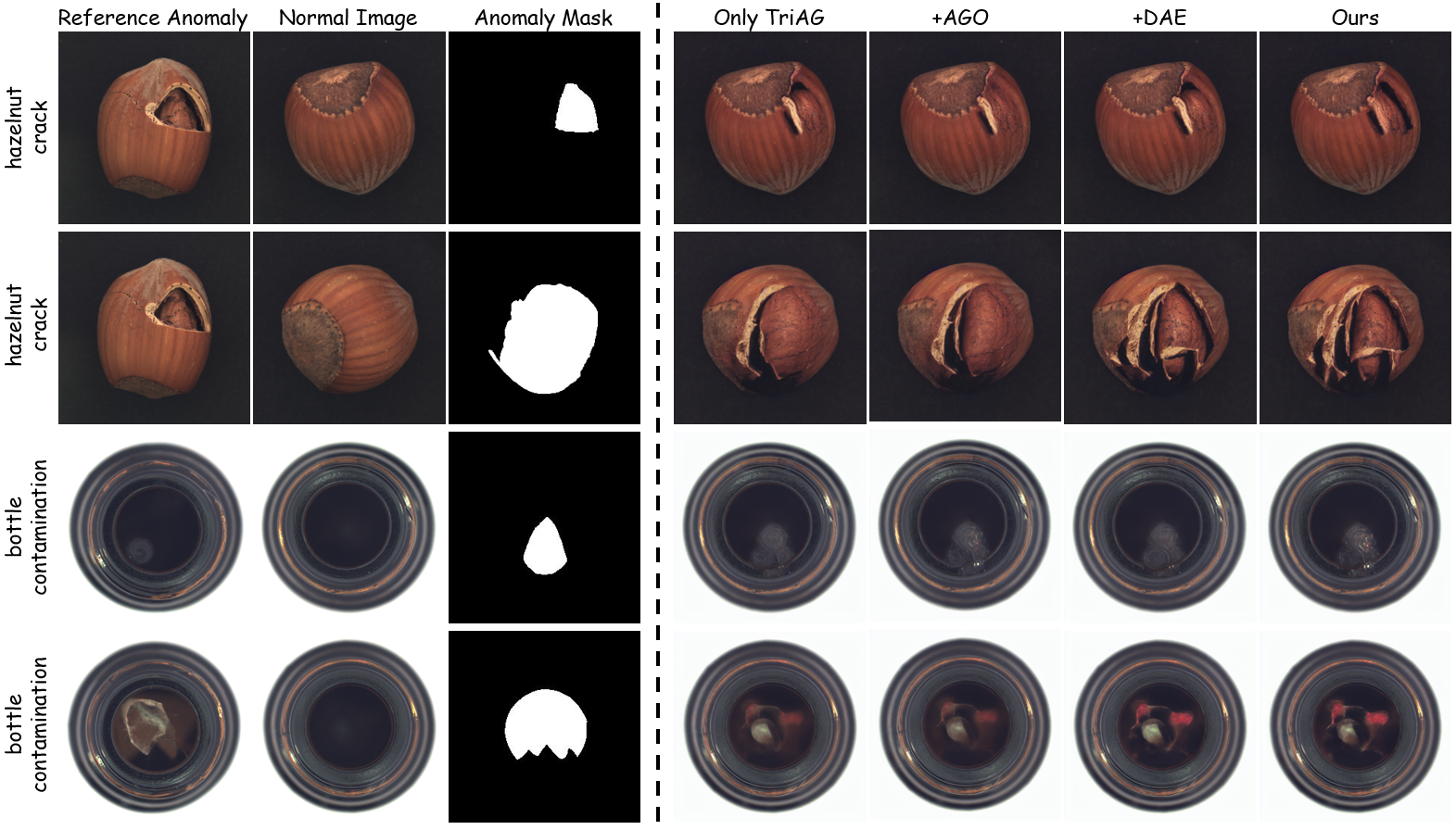}
  \caption{Qualitative ablation of TriAG, AGO, and DAE for anomaly realism and localization.}
  \label{fig:app-22}
\end{figure*}

We ablate the three components of \method—TriAG, DAE, and AGO. 
TriAG is kept active as the backbone in all variants, while DAE and AGO are removed individually and jointly to isolate their contributions. Additional anomaly-generation results are provided in Fig.~\ref{fig:app-22}.

Images to the left of the divider show the reference anomalous image, the normal image, and the target anomaly mask used for synthesizing new anomalies. In the \textit{hazelnut–hole} case, our method can use one same reference anomaly image to guide the synthesis of more realistic defects. The visualization results indicate that the TriAG backbone effectively captures realistic anomaly semantics, but the synthesized anomalies may do not fully occupy the target anomaly mask, which limits improvements in downstream anomaly localization. To address this issue, we introduce the DAE module to enhance the visibility of anomalies. AGO further provides text-level guidance for anomaly synthesis, injecting more realistic anomalous textures and benefiting downstream classification performance. Finally, by combining attention editing for image-level anomaly semantics with text-level anomalous prompt optimization, our method can synthesize realistic and diverse anomalies, thereby improving downstream anomaly detection performance.


\textbf{Mask acquisition \& Impact.}
We also investigate the impact of mask sources on our synthesis process. Standard approaches generate masks via heuristics (e.g., Perlin noise in DRAEM) or learn mask distributions from few-shot samples. Aligning with these protocols, we utilize diverse mask sources for synthesis: specifically, AnomalyDiffusion masks for MVTec-AD and SeaS for VisA/Real-IAD.

Furthermore, we evaluate the effect of four mask sources on the \textit{hazelnut-crack} category. For the Perlin noise setting, we randomly generate masks following the DRAEM approach and subsequently compute the intersection with the foreground mask of the normal image. The Nano Banana setting utilizes masks synthesized by Google's current image generation model. Experiments on \textit{hazelnut-crack} confirm our robustness against these diverse mask sources (Table~\ref{tab:mask-random}).

\begin{table}[h]
\centering
\caption{Robustness to different mask source for our method.}
\label{tab:mask-random}
\vspace{-0.2cm} 

\setlength{\tabcolsep}{4.5pt}
\renewcommand{\arraystretch}{0.72} 
\resizebox{0.98\linewidth}{!}{
\begin{tabular}{c|ccc|cccc}
\toprule
Mask Source & AUROC-I & AP-I & F1-I & AUROC-P & AP-P & F1-P & Pro \\[-0.5ex]
\midrule
Perlin noise  & 100 & 100 & 100 & 99.8 & 95.5 & 89.8 & 94.0 \\
Nano Banana & 100 & 100 & 100 & 99.9 & 97.3 & 92.1 & 95.1 \\
AnomalyDiffusion & 100 & 100 & 100 & 99.9 & 96.3 & 90.6 & 95.2 \\
SeaS    & 100 & 100 & 100 & 99.6 & 94.8 & 89.3 & 95.9 \\
\bottomrule
\end{tabular}
}
\vspace{-0.05cm}
\end{table}


\section{Experiments on VisA Dataset}
\label{app-F}
To validate the effectiveness of our approach, we additionally evaluate downstream anomaly detection and localization on VisA using anomalies synthesized by our method.

\noindent\textbf{VisA dataset.}
The VisA dataset comprises 12 object categories spanning three domains. Its anomalous images exhibit a broad spectrum of defects, ranging from surface imperfections—such as scratches, dents, stains, and cracks—to logical anomalies, including misalignment and missing parts.

\noindent\textbf{Experimental settings.}
We use VisA dataset divided by SeaS according to defect categories, while dividing the good data in the original dataset into train and good. We use SeaS to generate masks for anomaly synthesis, since AnomalyDiffusion often produces empty masks on VisA.


Following the MVTec-AD setup, we designate the first one-third images of VisA as reference images for anomaly synthesis and use the remaining two-thirds to evaluate downstream anomaly detection and localization.
We evaluate three training-based anomaly generation baselines on VisA—AnomalyDiffusion, DualAnoDiff, and SeaS. We also include the training-free AnomalyAny for comparison. For each method, we synthesize 1,000 image–mask pairs and combine them with the first one-third of real anomalous images to train a U-Net segmentation model under the same experimental protocol as in MVTec-AD.
Because detailed prompts for VisA and generation details in AnomalyAny are not publicly available, we instead report its best VisA detection performance from the paper under the \emph{full-shot} setting.\footnote{The full-shot setting in AnomalyAny uses all normal images and generates 3-5 anomalous images conditioned on each normal image for training detection model.}

\begin{figure*}[t]
  \centering
  \includegraphics[width=\textwidth]{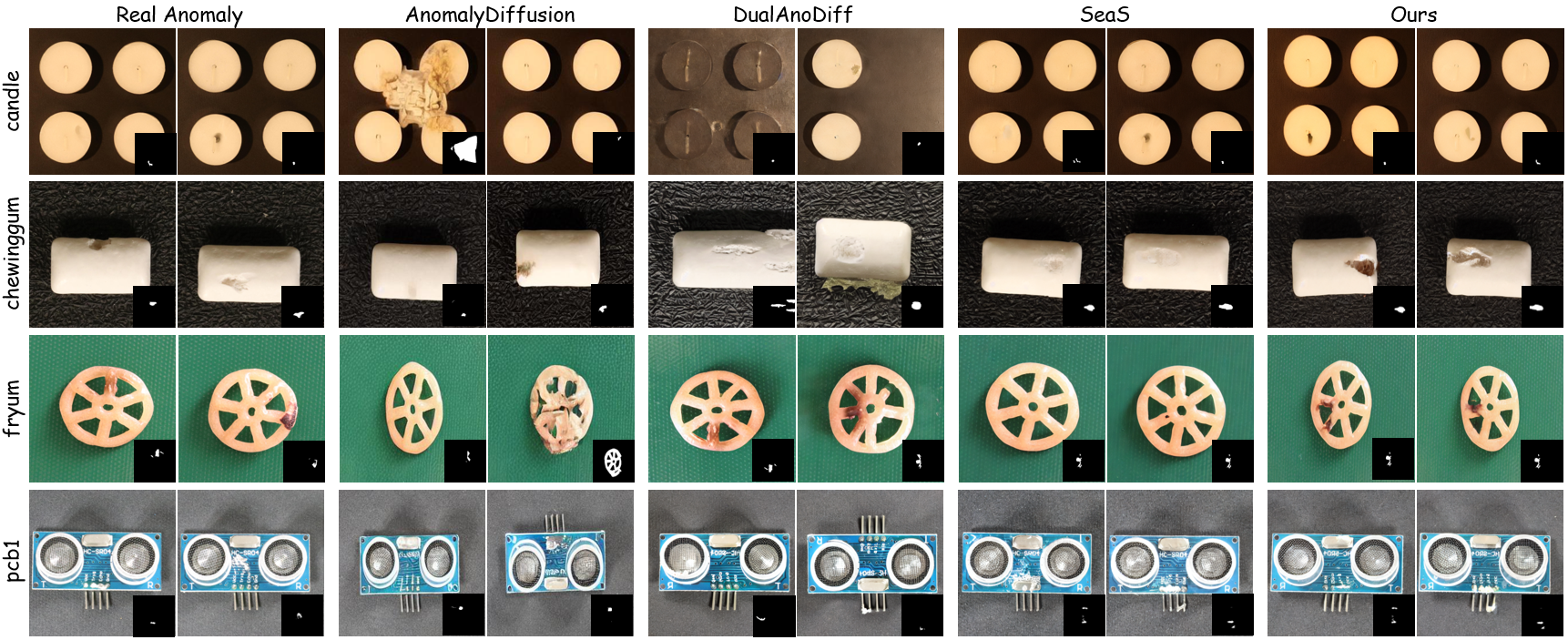}
  \caption{Qualitative comparison of generated results on VisA. The sub-image in the lower right corner is anomaly mask.}
  \label{fig:app-23}
\end{figure*}

\noindent\textbf{Anomaly image generation quality.}
Fig.~\ref{fig:app-23} presents qualitative results on representative VisA categories synthesized by different methods. Since AnomalyAny cannot generate precise anomaly masks, it is excluded from this comparison. As shown in Fig.~\ref{fig:app-23}, AnomalyDiffusion still fails to correctly synthesize small anomalies on VisA. DualAnoDiff produces anomalous samples that deviate from the real data distribution, often corrupting the background and generating unrealistic objects. Using the same masks from SeaS, we observe that SeaS does not consistently produce mask-filling anomalies within the masked regions, which limits the improvement on downstream detection. In contrast, our method not only generates realistic anomalies but also faithfully fills the anomaly masks.

\setlength{\textfloatsep}{6pt}
\setlength{\intextsep}{6pt}
\begin{table}[t]
\centering
\small
\setlength{\tabcolsep}{6pt}
\caption{Comparison of different methods on VisA. Bold and underlined text indicate the best and second-best results, respectively.}
\label{tab:icl_tryon_or_animation}
\resizebox{\columnwidth}{!}{
\begin{tabular}{c|cc|cc|cc|cc}
\toprule
\multirow{2}{*}{Category} &
\multicolumn{2}{c|}{AnomalyDiffusion~\cite{hu2024anomalydiffusion}} &
\multicolumn{2}{c|}{DualAnoDiff~\cite{jin2025dual}} &
\multicolumn{2}{c|}{SeaS~\cite{dai2025seasfewshotindustrialanomaly}} &
\multicolumn{2}{c}{Ours} \\
& KID$\downarrow$ & IC\!-L$\uparrow$
& KID$\downarrow$ & IC\!-L$\uparrow$
& KID$\downarrow$ & IC\!-L$\uparrow$
& KID$\downarrow$ & IC\!-L$\uparrow$ \\
\midrule
capsules      & \underline{103.91} & 0.56 & 126.19 & \textbf{0.64} & 111.32 & \underline{0.61} & \textbf{20.53} & 0.54 \\
candle        & 183.21 & \underline{0.17} & 180.22 & \textbf{0.38} & \underline{47.65} & 0.10 & \textbf{41.91} & 0.12 \\
cashew        & 78.22 & \underline{0.38} & 52.46 & \textbf{0.48} & \underline{10.37} & 0.35 & \textbf{10.29} & 0.35 \\
chewinggum    & 39.63 & \underline{0.34} & \underline{29.76} & \textbf{0.43} & \textbf{24.19} & 0.29 & 33.36 & 0.33 \\
fryum         & 238.41 & \underline{0.27} & 102.00 & \textbf{0.41} & \underline{82.20} & 0.24 & \textbf{24.93} & 0.18 \\
macaroni1     & 153.18 & 0.22 & 207.27 & \textbf{0.40} & \underline{152.04} & \underline{0.22} & \textbf{44.92} & 0.19 \\
macaroni2     & 178.19 & 0.35 & 221.05 & \textbf{0.50} & \underline{150.64} & \underline{0.40} & \textbf{26.83} & 0.35 \\
pcb1          & 77.68 & \underline{0.31} & 50.66 & \textbf{0.40} & \textbf{29.91} & 0.31 & \underline{36.03} & 0.29 \\
pcb2          & 81.66 & 0.29 & 36.04 & \textbf{0.38} & \underline{30.80} & \underline{0.30} & \textbf{10.91} & 0.27 \\
pcb3          & \underline{52.12} & 0.23 & 81.88 & \textbf{0.35} & 83.62 & \underline{0.25} & \textbf{8.63} & 0.21 \\
pcb4          & 34.29 & \underline{0.29} & 28.68 & \textbf{0.37} & \underline{12.31} & 0.25 & \textbf{9.82} & 0.24 \\
pipe\_fryum   & 76.04 & 0.21 & 47.70 & \textbf{0.40} & \underline{23.59} & 0.19 & \textbf{9.02} & \underline{0.23} \\
\midrule[1.1pt]
\rowcolor{AvgRow}
\textbf{Average} & 108.04 & \underline{0.30} & 96.99 & \textbf{0.43} & \underline{63.22} & 0.29 & \textbf{23.10} & 0.27 \\
\bottomrule[1.1pt]
\end{tabular}}
\end{table}

\noindent\textbf{Anomaly generation for anomaly detection and localization.}
We report image-level and pixel-level anomaly detection results on VisA in Table~\ref{tab:img_level_visA} and Table~\ref{tab:pix_level_visA}, respectively. In addition, Table~\ref{tab:visA_any} compares our method with AnomalyAny. Since AnomalyAny does not report AP, we only compare AUROC and F1-max scores. Our method is slightly inferior to DualAnoDiff at the image level, as DualAnoDiff introduces a dedicated anomaly branch to learn foreground defects. Nevertheless, despite the challenges posed by small anomalies discussed in Sec.~\ref{limitations}, our approach still achieves the best performance at the pixel level.

\begin{table}[t]
\centering
\footnotesize
\setlength{\tabcolsep}{3.5pt}
\renewcommand{\arraystretch}{0.95}
\caption{Comparison to training-free AnomalyAny on VisA.}
\label{tab:visA_any}
\resizebox{\columnwidth}{!}{
\begin{tabular}{c|cc|cc|cc|cc}
\toprule
\multirow{2}{*}{Category} 
& \multicolumn{4}{c|}{AnomalyAny} & \multicolumn{4}{c}{Ours} \\
\cmidrule(lr){2-9}
& I-AUC & I-F1 & P-AUC & P-F1 & I-AUC & I-F1 & P-AUC & P-F1 \\
\midrule
candle      & 95.6 & 90.0 & 99.3 & 40.1 & 94.9 & 86.1 & 98.8 & 42.4 \\
capsules    & 96.2 & 93.8 & 99.1 & 60.1 & 89.6 & 83.2 & 99.2 & 68.7 \\
cashew      & 97.4 & 94.5 & 99.2 & 70.4 & 94.0 & 91.7 & 99.9 & 92.3 \\
chewinggum  & 98.7 & 97.0 & 99.5 & 75.3 & 98.5 & 94.8 & 99.7 & 78.1 \\
fryum       & 98.4 & 97.5 & 97.4 & 53.6 & 94.4 & 91.5 & 98.6 & 70.8 \\
macaroni1   & 95.3 & 88.5 & 99.5 & 36.4 & 99.1 & 95.5 & 99.9 & 49.7 \\
macaroni2   & 84.7 & 79.3 & 99.6 & 28.9 & 83.4 & 74.2 & 98.5 & 30.9 \\
pcb1        & 95.9 & 91.8 & 98.8 & 41.8 & 93.7 & 82.9 & 99.7 & 83.4 \\
pcb2        & 94.1 & 88.2 & 98.2 & 40.3 & 96.3 & 89.8 & 94.8 & 53.9 \\
pcb3        & 95.9 & 90.4 & 97.5 & 52.9 & 97.1 & 92.9 & 98.2 & 64.6 \\
pcb4        & 99.4 & 96.6 & 98.4 & 46.5 & 98.7 & 95.0 & 99.1 & 62.5 \\
pipe\_fryum & 98.4 & 95.9 & 99.1 & 58.7 & 83.5 & 82.7 & 99.8 & 78.2 \\
\midrule
\rowcolor{AvgRow}
\textbf{Average} & \textbf{95.8} & \textbf{91.9} & 98.7 & 50.4 & 93.6 & 88.4 & \textbf{98.9} & \textbf{64.6} \\
\bottomrule
\end{tabular}}
\end{table}

\begin{figure*}[t]
  \centering
  \includegraphics[width=\textwidth]{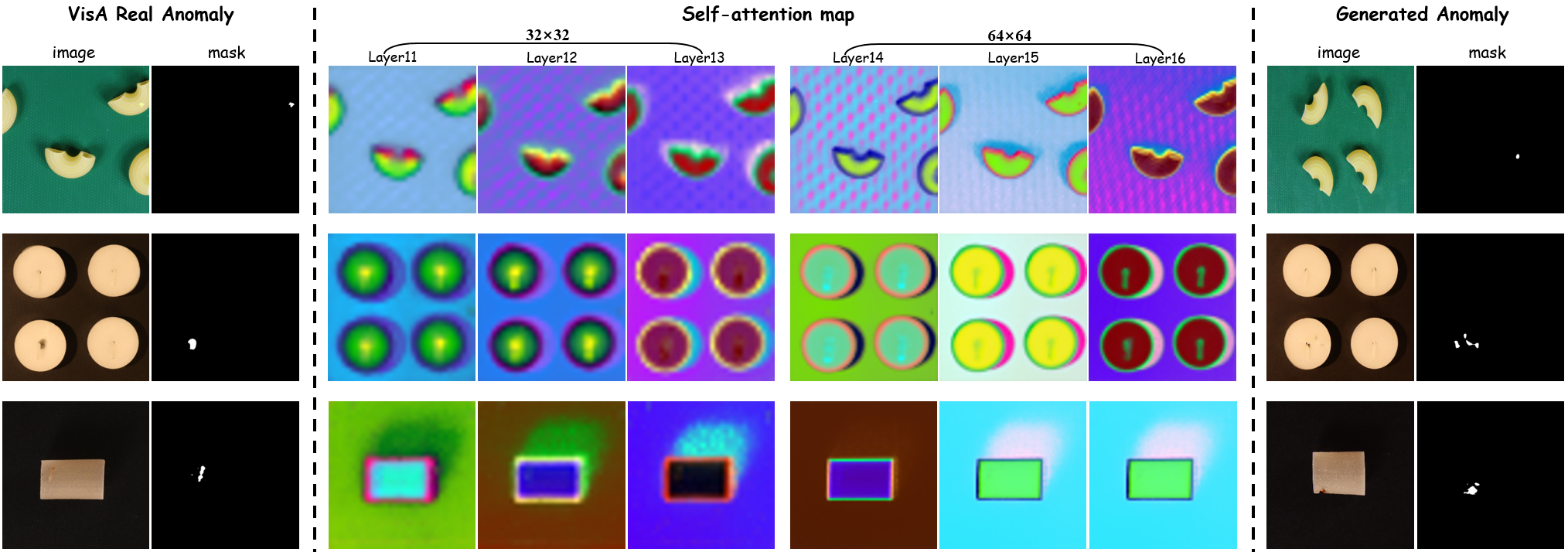}
  \caption{Limitations of Self-Attention for Tiny Anomaly Generation. The middle columns show the three leading principal components of the self-attention matrix for the real reference anomaly on the left at the 30th denoising step, while the right columns present our synthesized anomalous images and the corresponding anomaly masks. For small anomalous regions, the anomaly is almost invisible in the self-attention feature maps spanned by the top three PCA components, which makes it challenging to synthesize anomalies within small masks.}
  \label{fig:app-1}
\end{figure*}

\section{Experiments on Real-IAD Dataset}
\label{app-I}
Additionally, we evaluate our method's robustness under the one-reference setting using the Real-IAD dataset. This large-scale dataset provides five-view images encompassing 30 categories and 111 anomaly types, with a total of over 150K images. Each type roughly has 120 anomalous images. We use the first image from the top view as the reference and employ SeaS masks to control for spatial variation. We report both downstream AD performance using a trained U-Net in Table~\ref{tab:one-shot} and resource usage in Table~\ref{tab:time_speed}.
Training Time is computed on a single 48GB NVIDIA RTX 5880 GPU. Measured on a single NVIDIA RTX 5880 GPU, training-based methods incur high cumulative costs by requiring separate models per category (i.e., 30 distinct models for Real-IAD). Conversely, our training-free method outperforms them, relying on a frozen SD model to generalize across datasets.

\begin{table}[h]
\centering
\caption{One-reference AD results on Real-IAD (trained U-Net).}
\label{tab:one-shot}
\vspace{-0.20cm} 
\small
\setlength{\tabcolsep}{4.5pt}
\renewcommand{\arraystretch}{0.72} 
\resizebox{0.98\linewidth}{!}{
\begin{tabular}{c|ccc|cccc}
\toprule
Method & AUROC-I & AP-I & F1-I & AUROC-P & AP-P & F1-P & Pro \\[-0.5ex]
\midrule
DualAnoDiff & 72.5 & 77.8 & 75.8 & 91.3 & 25.7 & 31.9 & 75.7 \\
SeaS        & 76.7 & 80.4 & 77.2 & 87.7 & 31.1 & 36.5 & 76.9 \\
Ours        & \textbf{79.3} & \textbf{84.0} & \textbf{80.4} & \textbf{93.6} & \textbf{37.3} & \textbf{41.9} & \textbf{82.9} \\[-0.4ex]
\bottomrule
\end{tabular}
}
\vspace{-0.1cm} 
\end{table}

\begin{table}[h]
\centering
\small
\setlength{\tabcolsep}{8pt}
\caption{Comparison of runtime and memory cost.}
\label{tab:time_speed}
\vspace{-0.20cm}
\resizebox{\columnwidth}{!}{
\renewcommand{\arraystretch}{0.8}
\begin{tabular}{c|c|c|c}
\toprule
\textbf{Methods} & \textbf{Training Time (single GPU)} & \textbf{Inference Time (per image)} & \textbf{GPU Memory} \\
\midrule
DualAnoDiff                 & 119 hours & 23\,s & 12G \\
SeaS & 160 hours & 11\,s & 24G \\
\midrule
\method\ Ours                                & 0 hours        & 28\,s   & 32G \\
\bottomrule
\end{tabular}}
\vspace{-0.05cm}
\end{table}

\section{Additional Results}
\label{app-G}
\subsection{More anomaly detection and localization Results}
Tables~\ref{tab:mvtec_image} and \ref{tab:mvtec_pixel} report detailed image-level and pixel-level metrics on MVTec-AD for each category, where we first generate 1,000 anomalous images per category and then combine them with the first one-third of real anomalous images in MVTec-AD dataset to train a U-Net segmentation model for downstream anomaly detection and localization. Our method achieves the best performance on both image-level and pixel-level anomaly detection tasks.


\begin{table*}[t]
\centering
\small
\setlength{\tabcolsep}{7pt}
\caption{Comparison on a trained U-Net segmentation model for \textbf{image-level} anomaly detection and localization on \textbf{VisA} dataset.}
\label{tab:img_level_visA}
\resizebox{\textwidth}{!}{
\begin{tabular}{l|ccc|ccc|ccc|ccc}
\toprule
\multirow{2}{*}{Category} &
\multicolumn{3}{c|}{AnomalyDiffusion} &
\multicolumn{3}{c|}{DualAnoDiff} &
\multicolumn{3}{c|}{SeaS} &
\multicolumn{3}{c}{Ours} \\
\cmidrule(lr){2-4}\cmidrule(lr){5-7}\cmidrule(lr){8-10}\cmidrule(lr){11-13}
& AUC-I & AP-I & F1-I & AUC-I & AP-I & F1-I & AUC-I & AP-I & F1-I & AUC-I & AP-I & F1-I \\
\midrule
candle       & \underline{95.1} & \underline{95.3} & \underline{90.1} & \textbf{96.7} & \textbf{95.6} & \textbf{90.4} & 78.7 & 77.9 & 68.5 & 94.9 & 93.3 & 86.1 \\
capsules     & \textbf{91.9} & \textbf{93.8} & \textbf{85.9} & \underline{91.0} & \underline{93.0} & \underline{85.7} & 84.7 & 89.0 & 78.6 & 89.6 & 91.9 & 83.2 \\
cashew       & \underline{95.2} & \underline{96.0} & 92.2 & \textbf{99.2} & \textbf{99.4} & \textbf{96.4} & 94.8 & 95.2 & \underline{92.4} & 94.0 & 94.6 & 91.7 \\
chewinggum   & 94.4 & 97.2 & 93.1 & \textbf{99.6} & \textbf{99.7} & \textbf{97.8} & \underline{98.9} & \underline{99.3} & \underline{95.0} & 98.5 & 99.0 & 94.8 \\
fryum        & 81.3 & 87.2 & 79.2 & 86.7 & 90.7 & 84.1 & \underline{91.2} & \underline{94.4} & \underline{86.7} & \textbf{94.4} & \textbf{96.7} & \textbf{91.5} \\
macaroni1    & 94.6 & 92.7 & 86.5 & 96.8 & 95.6 & 90.3 & \underline{98.8} & \underline{98.4} & \underline{92.9} & \textbf{99.1} & \textbf{98.9} & \textbf{95.5} \\
macaroni2    & 67.3 & 48.9 & 41.0 & \textbf{91.1} & \textbf{89.9} & \textbf{78.3} & \underline{86.3} & \underline{86.0} & \underline{77.6} & 83.4 & 83.1 & 74.2 \\
pcb1         & 90.9 & 88.4 & 79.5 & \textbf{96.2} & \textbf{96.0} & \underline{90.6} & \underline{95.7} & \underline{95.8} & \textbf{92.1} & 93.7 & 91.9 & 82.9 \\
pcb2         & 94.8 & 92.9 & 88.3 & \underline{95.8} & \underline{95.8} & \textbf{90.4} & 94.5 & 94.5 & 88.1 & \textbf{96.3} & \textbf{96.0} & \underline{89.8} \\
pcb3         & 92.8 & 90.8 & 81.8 & 95.0 & 94.3 & 90.2 & \textbf{99.0} & \textbf{98.6} & \textbf{93.8} & \underline{97.1} & \underline{97.1} & \underline{92.9} \\
pcb4         & 94.8 & 92.1 & 82.3 & 98.1 & \textbf{98.4} & \underline{95.0} & \textbf{98.9} & 95.9 & \textbf{97.2} & \underline{98.7} & \underline{97.2} & \underline{95.0} \\
pipe\_fryum  & 88.3 & 91.2 & 85.5 & \textbf{95.2} & \textbf{96.9} & \textbf{90.1} & \underline{92.0} & \underline{94.3} & \underline{86.5} & 83.5 & 87.8 & 82.7 \\
\midrule
\rowcolor{AvgRow}
\textbf{Average} & 90.1 & 88.9 & 82.1 & \textbf{95.1} & \textbf{95.4} & \textbf{89.9} & 92.8 & 93.3 & 87.5 & \underline{93.6} & \underline{94.0} & \underline{88.4} \\
\bottomrule
\end{tabular}}
\end{table*}

{
\setlength{\abovecaptionskip}{4pt}
\setlength{\belowcaptionskip}{-2pt}
\begin{table*}[t]
\setlength{\textfloatsep}{5pt}
\centering
\small
\setlength{\tabcolsep}{7pt}
\caption{Comparison on a trained U-Net segmentation model for \textbf{pixel-level} anomaly detection and localization on \textbf{VisA} dataset.}
\label{tab:pix_level_visA}
\resizebox{\textwidth}{!}{
\begin{tabular}{l|ccc|ccc|ccc|ccc}
\toprule
\multirow{2}{*}{Category} &
\multicolumn{3}{c|}{AnomalyDiffusion} &
\multicolumn{3}{c|}{DualAnoDiff} &
\multicolumn{3}{c|}{SeaS} &
\multicolumn{3}{c}{Ours} \\
\cmidrule(lr){2-4}\cmidrule(lr){5-7}\cmidrule(lr){8-10}\cmidrule(lr){11-13}
& AUC-P & AP-P & F1-P & AUC-P & AP-P & F1-P & AUC-P & AP-P & F1-P & AUC-P & AP-P & F1-P \\
\midrule
candle       & \textbf{99.2} & 24.7 & 33.0 & \underline{98.9} & \textbf{48.8} & \textbf{49.7} & 96.5 & 29.9 & 37.5 & 98.8 & \underline{39.8} & \underline{42.4} \\
capsules     & \textbf{99.6} & 60.2 & 61.2 & 98.4 & \underline{60.7} & \underline{63.8} & \underline{99.5} & 59.1 & 62.3 & 99.2 & \textbf{70.2} & \textbf{68.7} \\
cashew       & \underline{98.4} & 68.8 & 63.9 & \textbf{99.9} & \underline{98.0} & \underline{94.5} & \textbf{99.9} & \textbf{98.2} & \textbf{95.3} & \textbf{99.9} & 96.4 & 92.3 \\
chewinggum   & 98.7 & 78.3 & 72.4 & \textbf{99.9} & \textbf{86.6} & \textbf{78.2} & 99.5 & 84.3 & 77.6 & \underline{99.7} & \underline{85.5} & \underline{78.1} \\
fryum        & 91.0 & 28.2 & 31.8 & \underline{97.9} & \underline{76.7} & \textbf{74.7} & 97.4 & 74.5 & 69.5 & \textbf{98.6} & \textbf{80.6} & \underline{70.8} \\
macaroni1    & 98.2 & 4.2  & 10.6 & \underline{99.8} & 28.7 & 37.6 & \textbf{99.9} & \textbf{50.4} & \textbf{53.3} & \textbf{99.9} & \underline{47.1} & \underline{49.7} \\
macaroni2    & 90.6 & 0.1  & 0.0  & 97.2 & \underline{19.7} & \underline{30.3} & \underline{97.3} & 17.8 & 26.7 & \textbf{98.5} & \textbf{22.4} & \textbf{30.9} \\
pcb1         & 99.0 & 76.6 & 75.8 & 98.9 & \underline{92.6} & \underline{88.1} & \underline{99.4} & \textbf{94.2} & \textbf{90.3} & \textbf{99.7} & 88.9 & 83.4 \\
pcb2         & \underline{98.2} & 23.1 & 37.1 & 96.1 & 35.8 & 46.6 & \textbf{99.1} & \underline{49.7} & \underline{49.3} & 94.8 & \textbf{51.3} & \textbf{53.9} \\
pcb3         & 93.6 & 36.7 & 43.6 & \underline{98.6} & 56.3 & 58.2 & \textbf{98.8} & \underline{68.6} & \underline{62.2} & 98.2 & \textbf{69.0} & \textbf{64.6} \\
pcb4         & 98.2 & 44.3 & 53.3 & \underline{98.8} & 52.3 & 53.3 & 98.6 & \underline{61.9} & \underline{60.5} & \textbf{99.1} & \textbf{66.1} & \textbf{62.5} \\
pipe\_fryum  & 99.0 & 69.9 & 66.7 & 99.4 & \underline{79.1} & 69.6 & \textbf{99.9} & \textbf{93.1} & \textbf{85.1} & \underline{99.8} & \underline{88.6} & \underline{78.2} \\
\midrule
\rowcolor{AvgRow}
\textbf{Average} & 97.0 & 42.9 & 45.8 & 98.7 & 61.3 & 62.1 & \underline{98.8} & \underline{65.1} & \underline{64.1} & \textbf{98.9} & \textbf{67.2} & \textbf{64.6} \\
\bottomrule
\end{tabular}}
\end{table*}
}

\begin{table*}[t]
\centering
\small
\caption{Comparison on a trained U-Net segmentation model for \textbf{image-level} anomaly detection and localization on \textbf{MVTec-AD} dataset.}
\label{tab:mvtec_image}
\setlength{\tabcolsep}{4pt}
\resizebox{\textwidth}{!}{
\begin{tabular}{l
|ccc|ccc|ccc|ccc|ccc}
\toprule
\multirow{2}{*}{Category}
& \multicolumn{3}{c|}{\textbf{DFMGAN}}
& \multicolumn{3}{c|}{\textbf{AnomalyDiffusion}}
& \multicolumn{3}{c|}{\textbf{DualAnoDiff}}
& \multicolumn{3}{c|}{\textbf{SeaS}}
& \multicolumn{3}{c}{\textbf{Ours}} \\
\cmidrule(lr){2-4}\cmidrule(lr){5-7}\cmidrule(lr){8-10}\cmidrule(lr){11-13}\cmidrule(lr){14-16}
& AUC-I & AP-I & F1-I
& AUC-I & AP-I & F1-I
& AUC-I & AP-I & F1-I
& AUC-I & AP-I & F1-I
& AUC-I & AP-I & F1-I \\
\midrule
bottle      & 99.3 & 99.8 & 97.7  & 99.8 & \underline{99.9} & \underline{98.9}  & \textbf{100} & \textbf{100} & \textbf{100} & \underline{99.9} & \underline{99.9} & \underline{98.9}  & \textbf{100} & \textbf{100} & \textbf{100} \\
cable       & 95.9 & 97.8 & 93.8  & \textbf{100} & \textbf{100} & \textbf{100}   & 99.8 & \underline{99.8} & 98.5 & \underline{98.0} & 98.8 & \underline{96.1}  & \textbf{100} & \textbf{100} & \textbf{100} \\
capsule     & 92.8 & 98.5 & 94.5  & \textbf{99.7} & \textbf{99.9} & \textbf{98.7}  & 96.3 & 99.2 & 94.7 & 97.1 & 99.2 & 95.4  & \underline{99.3} & \underline{99.8} & \underline{98.0} \\
carpet      & 67.9 & 87.9 & 87.3  & 96.7 & 98.8 & 94.3  & \underline{98.6} & \underline{99.9} & \underline{96.7} & 97.4 & 99.0 & \underline{96.7}  & \textbf{100} & \textbf{100} & \textbf{100} \\
grid        & 73.0 & 90.4 & 85.4  & 98.4 & 99.5 & 98.7  & \textbf{100} & 99.7 & \textbf{100} & 99.9 & \underline{99.9} & \underline{98.8}  & \textbf{100} & \textbf{100} & \textbf{100} \\
hazelnut    & \textbf{99.9} & \textbf{100} & \textbf{99.0}  & \underline{99.8} & \underline{99.9} & \underline{98.9}  & \textbf{99.9} & \textbf{100} & \textbf{99.0} & \underline{99.8} & 99.8 & \textbf{99.0}  & \textbf{99.9} & \textbf{100} & \textbf{99.0} \\
leather     & \underline{99.9} & \textbf{100} & \underline{99.2}  & \textbf{100} & \textbf{100} & \textbf{100}   & \textbf{100} & \textbf{100} & \textbf{100} & \textbf{100} & \textbf{100} & \textbf{100}   & \textbf{100} & \textbf{100} & \underline{99.2} \\
metal\_nut  & 99.3 & 99.8 & \underline{99.2}  & \textbf{100} & \textbf{100} & \textbf{100}   & \textbf{100} & \underline{99.9} & \textbf{100} & \underline{99.9} & \textbf{100} & \underline{99.2}  & \textbf{100} & \textbf{100} & \textbf{100} \\
pill        & 68.7 & 91.7 & 91.4  & 98.0 & \underline{99.6} & 97.0  & 98.2 & 99.0 & 95.9 & \underline{98.4} & \underline{99.6} & \textbf{97.9}  & \textbf{98.9} & \textbf{99.7} & \underline{97.5} \\
screw       & 22.3 & 64.7 & 85.3  & \textbf{96.8} & 97.9 & \textbf{95.5}  & 91.4 & 95.0 & 88.6 & 95.2 & \underline{98.0} & 92.5  & \underline{95.7} & \textbf{98.2} & \underline{94.9} \\
tile        & \textbf{100} & \textbf{100} & \textbf{100}   & \textbf{100} & \textbf{100} & \textbf{100}   & \underline{99.5} & 100 & \underline{98.3} & \textbf{100} & \textbf{100} & \textbf{100}   & \textbf{100} & \textbf{100} & \textbf{100} \\
toothbrush  & \textbf{100} & \textbf{100} & \textbf{100}   & \textbf{100} & \textbf{100} & \textbf{100}   & \textbf{100} & \textbf{100} & \textbf{100} & \textbf{100} & \textbf{100} & \textbf{100}   & \textbf{100} & \textbf{100} & \textbf{100} \\
transistor  & 90.8 & 92.5 & 88.9  & \textbf{100} & \textbf{100} & \textbf{100}   & \textbf{100} & \underline{99.7} & \textbf{100} & \underline{99.8} & \underline{99.5} & \underline{96.4}  & \textbf{100} & \textbf{100.0} & \textbf{100} \\
wood        & \underline{98.4} & \underline{99.4} & \textbf{98.8}  & \underline{98.4} & \underline{99.4} & \textbf{98.8}   & \underline{99.6} & \textbf{99.9} & \textbf{98.8} & 99.0 & 99.6 & \textbf{98.8}  & \textbf{99.7} & \underline{99.8} & \textbf{98.8} \\
zipper      & 99.7 & \underline{99.9} & \underline{99.4}  & \underline{99.9} & \textbf{100} & \underline{99.4}   & \textbf{100} & \textbf{100} & \textbf{100} & \textbf{100} & \textbf{100} & \textbf{100}   & \textbf{100} & \textbf{100} & \textbf{100} \\
\midrule
\rowcolor{AvgRow}
\textbf{Average}
& 87.2 & 94.8 & 94.7
& \underline{99.2} & \underline{99.7} & \underline{98.7}
& 98.9 & 98.9 & 98.0
& 99.0 & 99.6 & 98.0
& \textbf{99.6} & \textbf{99.8} & \textbf{99.2} \\
\bottomrule
\end{tabular}}
\end{table*}

\begin{table*}[t]
\centering
\small
\caption{Comparison on a trained U-Net segmentation model for \textbf{pixel-level} anomaly detection and localization on \textbf{MVTec-AD} dataset.}
\label{tab:mvtec_pixel}
\setlength{\tabcolsep}{4pt}
\resizebox{\textwidth}{!}{
\begin{tabular}{l
|ccc|ccc|ccc|ccc|ccc}
\toprule
\multirow{2}{*}{Category}
& \multicolumn{3}{c|}{\textbf{DFMGAN}}
& \multicolumn{3}{c|}{\textbf{AnomalyDiffusion}}
& \multicolumn{3}{c|}{\textbf{DualAnoDiff}}
& \multicolumn{3}{c|}{\textbf{SeaS}}
& \multicolumn{3}{c}{\textbf{Ours}} \\
\cmidrule(lr){2-4}\cmidrule(lr){5-7}\cmidrule(lr){8-10}\cmidrule(lr){11-13}\cmidrule(lr){14-16}
& AUC-P & AP-P & F1-P
& AUC-P & AP-P & F1-P
& AUC-P & AP-P & F1-P
& AUC-P & AP-P & F1-P
& AUC-P & AP-P & F1-P \\
\midrule
bottle      & 98.9 & 90.2 & 83.9 & 99.4 & 94.1 & 87.3 & \underline{99.5} & 93.4 & 85.7 & \textbf{99.7} & \textbf{95.9} & \textbf{88.8} & \textbf{99.7} & \underline{95.4} & \underline{88.4} \\
cable       & 97.2 & 81.0 & 75.4 & \underline{99.2} & \underline{90.8} & \underline{83.5} & 98.5 & 82.6 & 76.9 & 96.0 & 83.1 & 77.7 & \textbf{99.4} & \textbf{91.2} & \textbf{85.0} \\
capsule     & 79.2 & 26.0 & 35.0 & \underline{98.8} & \underline{57.2} & \underline{59.8} & \textbf{99.5} & \textbf{73.2} & \textbf{67.0} & 93.7 & 41.9 & 47.3 & 97.0 & 60.6 & 59.0 \\
carpet      & 90.6 & 33.4 & 38.1 & 98.6 & 81.2 & 74.6 & \underline{99.4} & \textbf{89.1} & \textbf{80.2} & 99.3 & 86.4 & 78.1 & \textbf{99.5} & \underline{88.5} & \underline{80.0} \\
grid        & 75.2 & 14.3 & 20.5 & 98.3 & 52.9 & 54.6 & 98.5 & 57.2 & 54.9 & \textbf{99.7} & \underline{76.3} & \underline{70.0} & \underline{99.6} & \textbf{78.6} & \textbf{71.6} \\
hazelnut    & \underline{99.7} & 95.2 & 89.5 & \textbf{99.8} & \underline{96.5} & \underline{90.6} & \textbf{99.8} & \textbf{97.7} & \textbf{92.8} & 99.5 & 92.3 & 85.6 & \textbf{99.8} & 96.2 & 90.1 \\
leather     & 98.5 & 68.7 & 66.7 & 99.8 & 79.6 & 71.0 & \textbf{99.9} & \textbf{88.8} & \underline{78.8} & \underline{99.8} & 85.2 & 77.0 & 99.7 & \underline{88.0} & \textbf{79.7} \\
metal\_nut  & 99.3 & 98.1 & \underline{94.5} & \textbf{99.8} & \underline{98.7} & 94.0 & \underline{99.6} & 98.0 & 93.0 & \textbf{99.8} & \textbf{99.2} & \textbf{95.7} & \textbf{99.8} & \textbf{99.2} & \textbf{95.7} \\
pill        & 81.2 & 67.8 & 72.6 & \underline{99.7} & 93.9 & \textbf{90.8} & 99.6 & 95.8 & 89.2 & \textbf{99.9} & \textbf{97.1} & \underline{90.7} & \underline{99.7} & \underline{96.1} & 89.9 \\
screw       & 58.8 & 2.2  & 5.3  & 97.0 & 51.8 & 50.9 & 98.1 & 57.1 & 56.1 & \underline{98.5} & \underline{58.5} & \underline{57.2} & \textbf{99.4} & \textbf{68.2} & \textbf{64.4} \\
tile        & 99.5 & 97.1 & 91.6 & 99.2 & 93.6 & 86.2 & 99.7 & 97.1 & 91.0 & \underline{99.8} & \underline{97.9} & \underline{92.5} & \textbf{99.9} & \textbf{98.2} & \textbf{92.7} \\
toothbrush  & 96.4 & \underline{75.9} & \underline{72.6} & \textbf{99.2} & \textbf{76.5} & \textbf{73.4} & 98.2 & 68.3 & 68.6 & \underline{98.4} & 70.0 & 68.1 & 96.3 & 58.6 & 59.2 \\
transistor  & 96.2 & 81.2 & 77.0 & \underline{99.3} & \underline{92.6} & \underline{85.7} & 98.0 & 86.7 & 79.6 & 98.0 & 87.3 & 81.9 & \textbf{99.9} & \textbf{98.2} & \textbf{93.2} \\
wood        & 95.3 & 70.7 & 65.8 & \textbf{98.9} & 84.6 & 74.5 & 99.4 & \textbf{91.6} & \textbf{83.8} & 99.0 & 87.0 & 79.6 & \underline{99.4} & \underline{89.4} & \underline{81.1} \\
zipper      & 92.9 & 65.6 & 64.9 & \underline{99.6} & 86.0 & 79.2 & \underline{99.6} & \textbf{90.7} & \textbf{82.7} & \textbf{99.7} & 88.2 & \underline{81.6} & 99.5 & \underline{88.5} & \underline{82.0} \\
\midrule[1.1pt]
\rowcolor{AvgRow}
\textbf{Average} &
90.0 & 62.7 & 62.1 &
99.1 & 81.4 & 76.3 &
99.1 & 84.5 & 78.8 &
98.7 & 83.1 & 78.1 &
\textbf{99.2} & \textbf{86.3} & \textbf{80.8} \\
\bottomrule[1.1pt]
\end{tabular}}
\end{table*}

\subsection{More anomaly generation results}

We conduct a comprehensive qualitative comparison between our generated results and existing anomaly image synthesis methods. Since the GAN-based method DFMGAN yields relatively limited visual fidelity, we primarily compare against diffusion-based approaches, with additional visualizations provided in Figs.~\ref{fig:app-4}–\ref{fig:app-18}. The leftmost column shows real anomalous and normal images from the training set, the middle columns show anomalies synthesized by training-based methods (AnomalyDiffusion, DualAnoDiff, SeaS), and the rightmost column presents the training-free AnomalyAny and our \method. Because AnomalyAny does not release its official prompts, we adopt the same prompt template as ours, “\textit{A photo of a [cls] with a [anomaly\_type]}”, and use GPT-5 to expand them into detailed text prompts for generation.

From these visualizations, we observe that AnomalyDiffusion generally preserves the normal background appearance but sometimes fails to synthesize clearly visible defects. DualAnoDiff and SeaS produce more pronounced anomalies, yet these often deviate from the true data distribution, and SeaS further introduces synthesis artifacts, which in turn limit downstream anomaly classification. In contrast, training-free methods better preserve the appearance of normal regions. However, AnomalyAny lacks guidance from real anomalous images and cannot reliably capture rich anomaly semantics from text alone. As a result, it often fails to capture the spatial relationship between anomalies and object regions, and may either produce no visible defects or synthesize defects that deviate from the real anomaly distribution, thereby limiting downstream anomaly detection and classification. By comparison, our \method does not require fine-tuning Stable Diffusion for each object category and under a single shared parameter setting, our \method can synthesize realistic and diverse anomalies, substantially improving downstream anomaly detection accuracy.

\subsection{Generalization to Real-World Scenes}
\begin{figure}[h]
  \centering
  \includegraphics[width=0.98\linewidth]{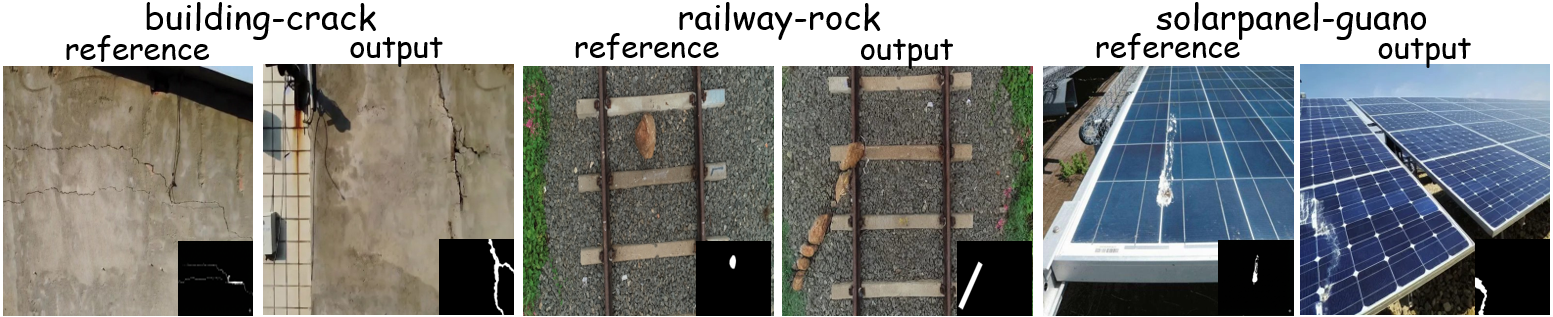}
  \vspace{-0.1cm}
  \caption{Generalization to real-world scenes.}
  \label{fig:real-world}
\end{figure}

Our method demonstrates robust generalization to real-world scenes (e.g., UAV imagery from UAV-RSOD dataset) beyond object-centric images by leveraging nano-banana generated masks (Fig.\ref{fig:real-world}). 

\section{Limitations}
\label{app-H}
\label{limitations}
Our method has two limitation: limit control on logical anomaly generation and self-attention grafting performs suboptimally when the reference anomaly is small.

\begin{figure}[h]
  \centering
  \includegraphics[width=0.98\linewidth]{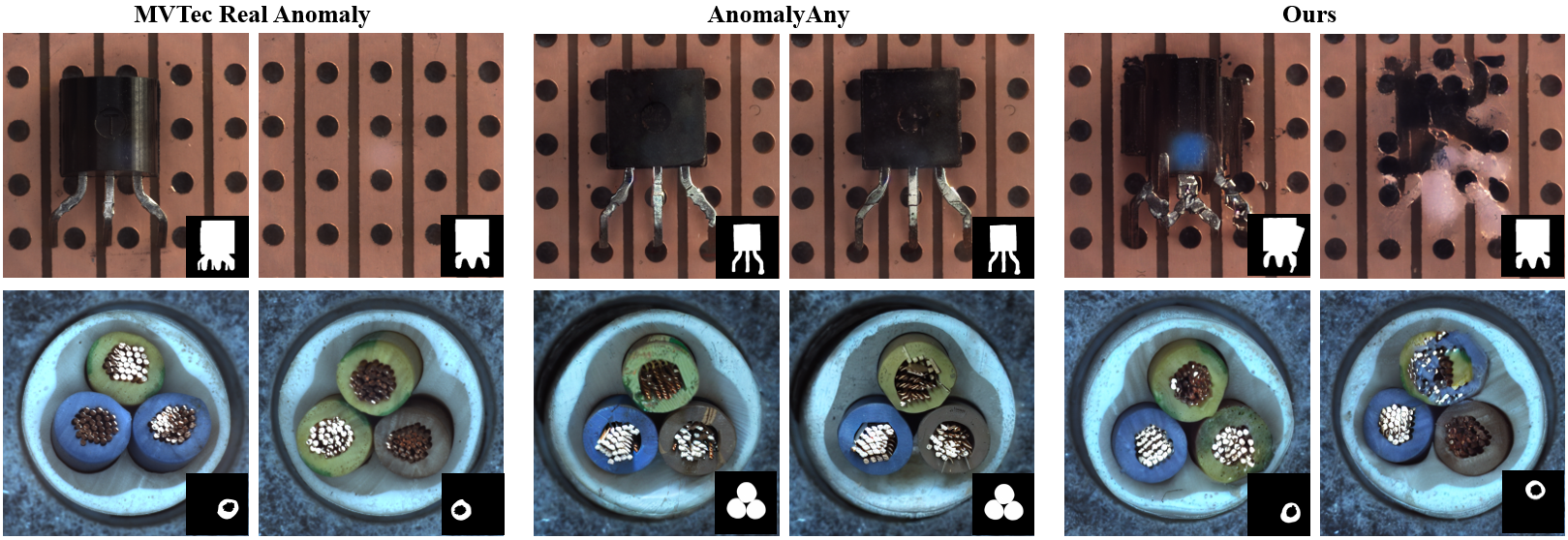}
  \vspace{-0.1cm}
  \caption{Logical anomaly generation.}
  \label{figa2:a3}
\end{figure}

\begin{figure*}[t]
  \centering
  \includegraphics[width=\textwidth]{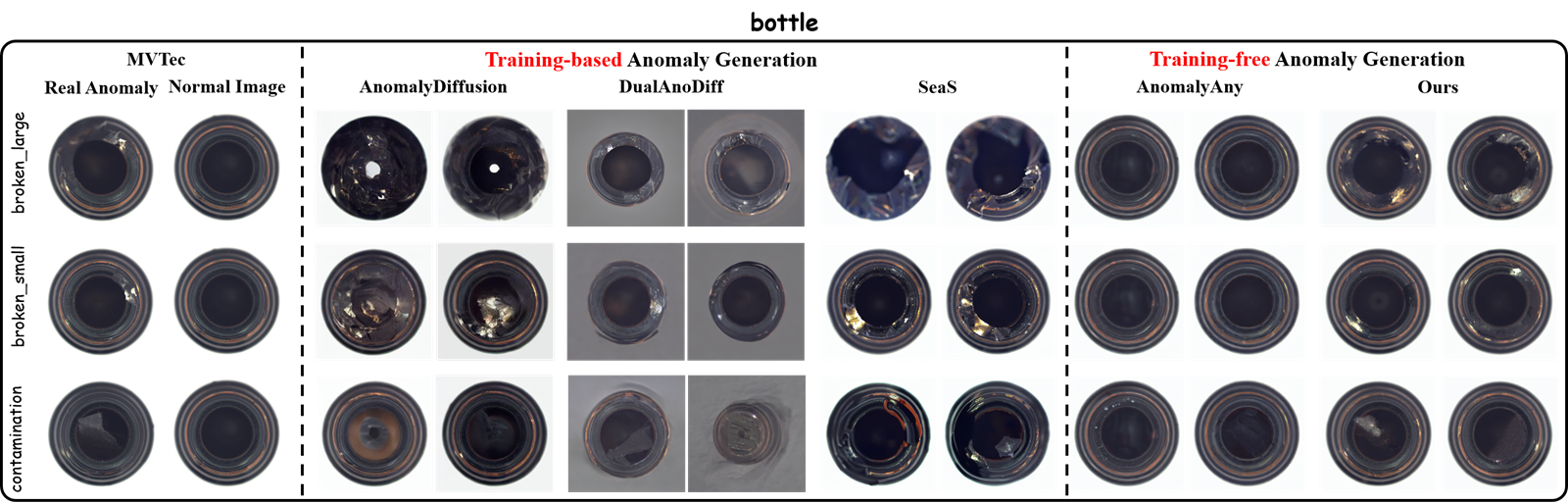}
  \caption{bottle qualitative results on MVTec-AD.}
  \label{fig:app-4}
\end{figure*}

\begin{figure*}[t]
  \centering
  \includegraphics[width=\textwidth]{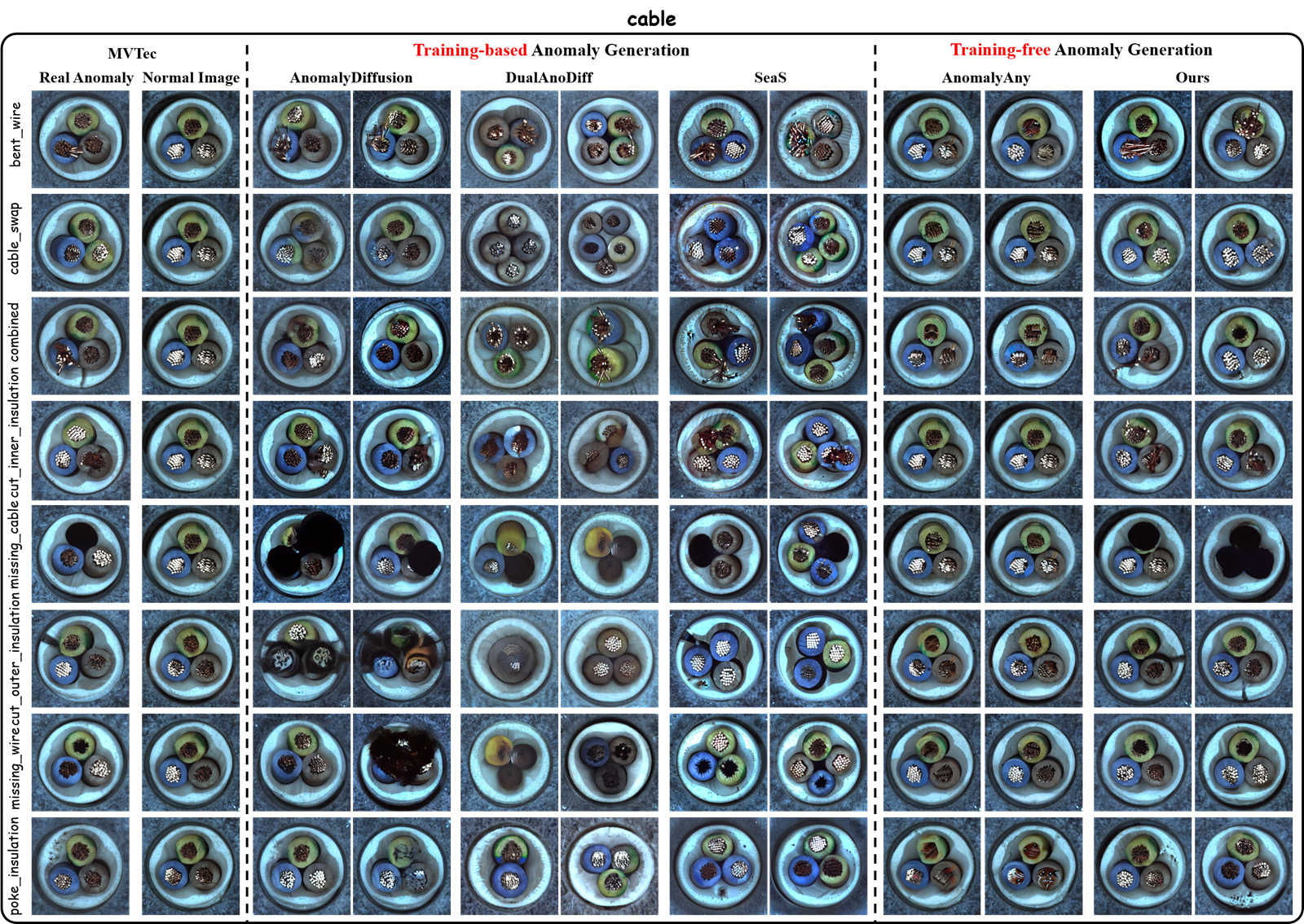}
  \caption{cable qualitative results on MVTec-AD.}
  \label{fig:app-5}
\end{figure*}

\begin{figure*}[t]
  \centering
  \includegraphics[width=\textwidth]{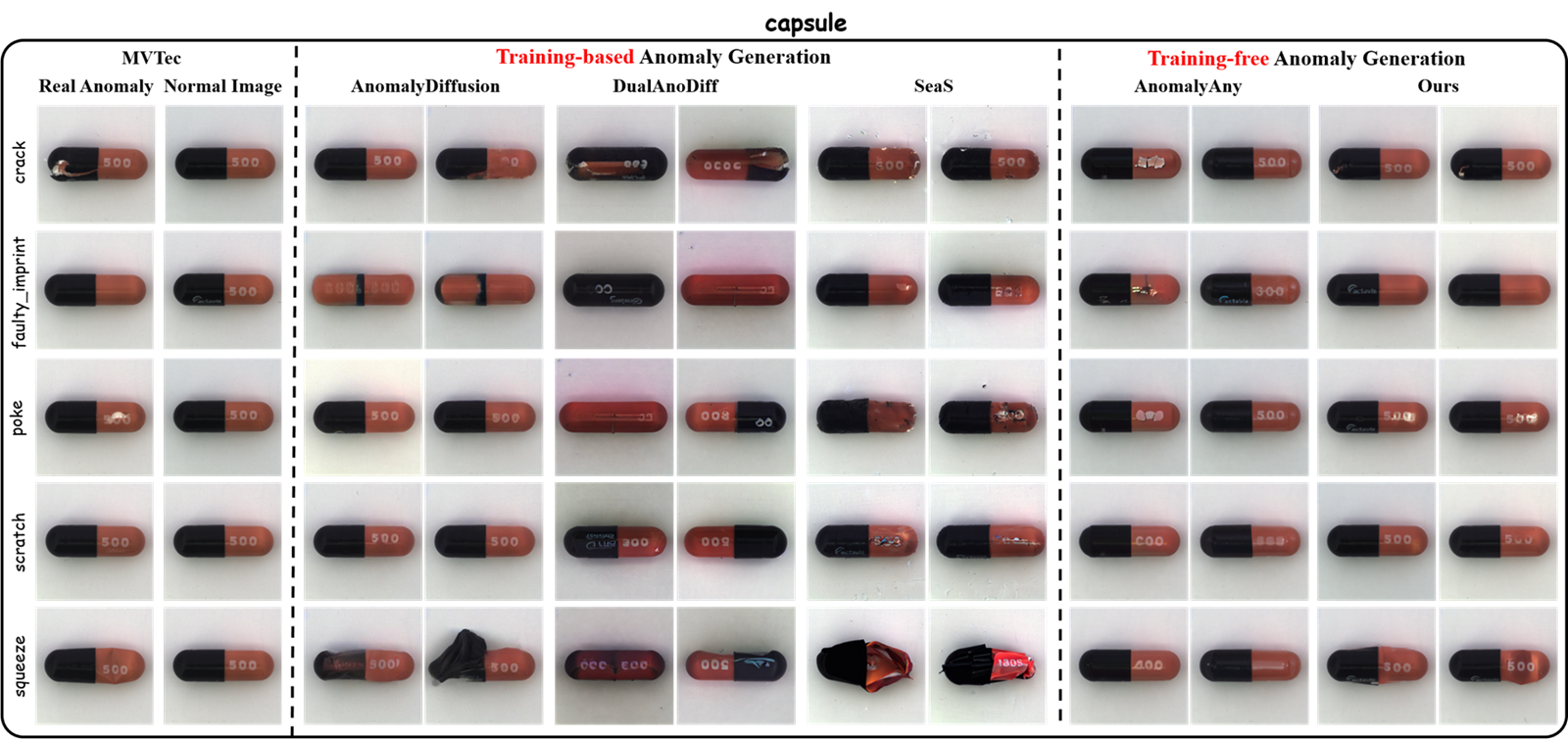}
  \caption{capsule qualitative results on MVTec-AD.}
  \label{fig:app-6}
\end{figure*}

\begin{figure*}[t]
  \centering
  \includegraphics[width=\textwidth]{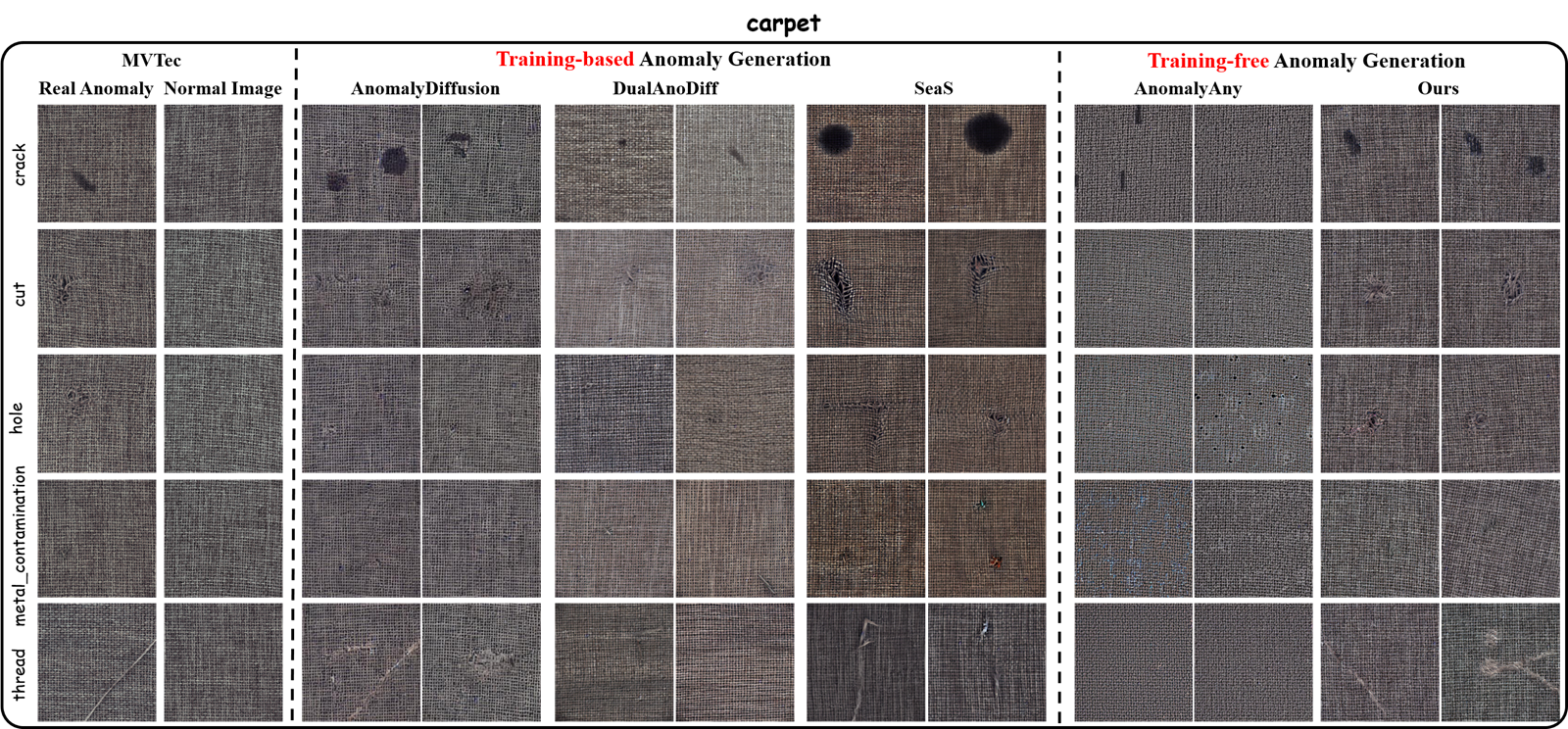}
  \caption{carpet qualitative results on MVTec-AD.}
  \label{fig:app-7}
\end{figure*}

\noindent\textbf{Limit control on logical anomaly generation.}
Our method synthesizes anomalies by manipulating attention and optimizing text embedding (AGO). However, compared with training-based approaches, it is less effective at reasoning aboutlogical anomalies. On MVTec-AD, two representative cases: cable-cable\_swap and transistor-misplaced, remain challenging.
As shown in Fig.~\ref{figa2:a3}, Transistor-misplaced typically denotes that the transistor pins are not correctly connected to the solder pads, or the transistor is missing. The training-free AnomalyAny enlarges attention on anomaly tokens. By using the same prompt template “\textit{A photo of a [cls] with a [anomaly\_type]}” and LLM-generated detailed prompts, it still fails to match the dataset distribution and does not produce the intended anomalies.
In our method, For misplacement case, the anomaly mask covers the transistor foreground. During self-attention grafting this can entangle features from the normal and reference branches, yielding inconsistent content as shown in Fig.~\ref{figa2:a3}. For the missing-transistor case, we do not manipulate attention in the initial five denoising steps, allowing the target branch to form an initial transistor silhouette that becomes difficult to remove later.
For cable-cable\_swap, among the expected green/blue/gray wires, two share the same color. Our approach further relies on precise mask–wire correspondence, so any misalignment of the mask often leads to bad generation.

We plan to integrate MLLMs to enrich prompt semantics for precise logical guidance in future work.

\noindent\textbf{Suboptimal on small anomaly generation.}
As discussed in \autoref{sec:Attention_Mechanism}, self-attention is applied to intermediate feature maps at \(64\times64\), \(32\times32\), \(16\times16\), and \(8\times8\). The reference anomaly image is accordingly downsampled to the corresponding spatial resolution. When the anomalous region is small, its representation in the self-attention space tends to exhibit unclear boundaries relative to the background. As illustrated
in Fig.~\ref{fig:app-1}, several VisA cases with small reference anomalies yield low-contrast self-attention maps. Concretely, at the 30th denoising step we apply principal component analysis (PCA) to the self-attention maps and visualize the top three principal components. Because the variance contributed by the anomalous region is much smaller than that of the background, the leading components tend to “ignore” the anomaly, which hampers synthesis of tiny defects.


\FloatBarrier

\begin{figure*}[t]
  \centering
  \includegraphics[width=\textwidth]{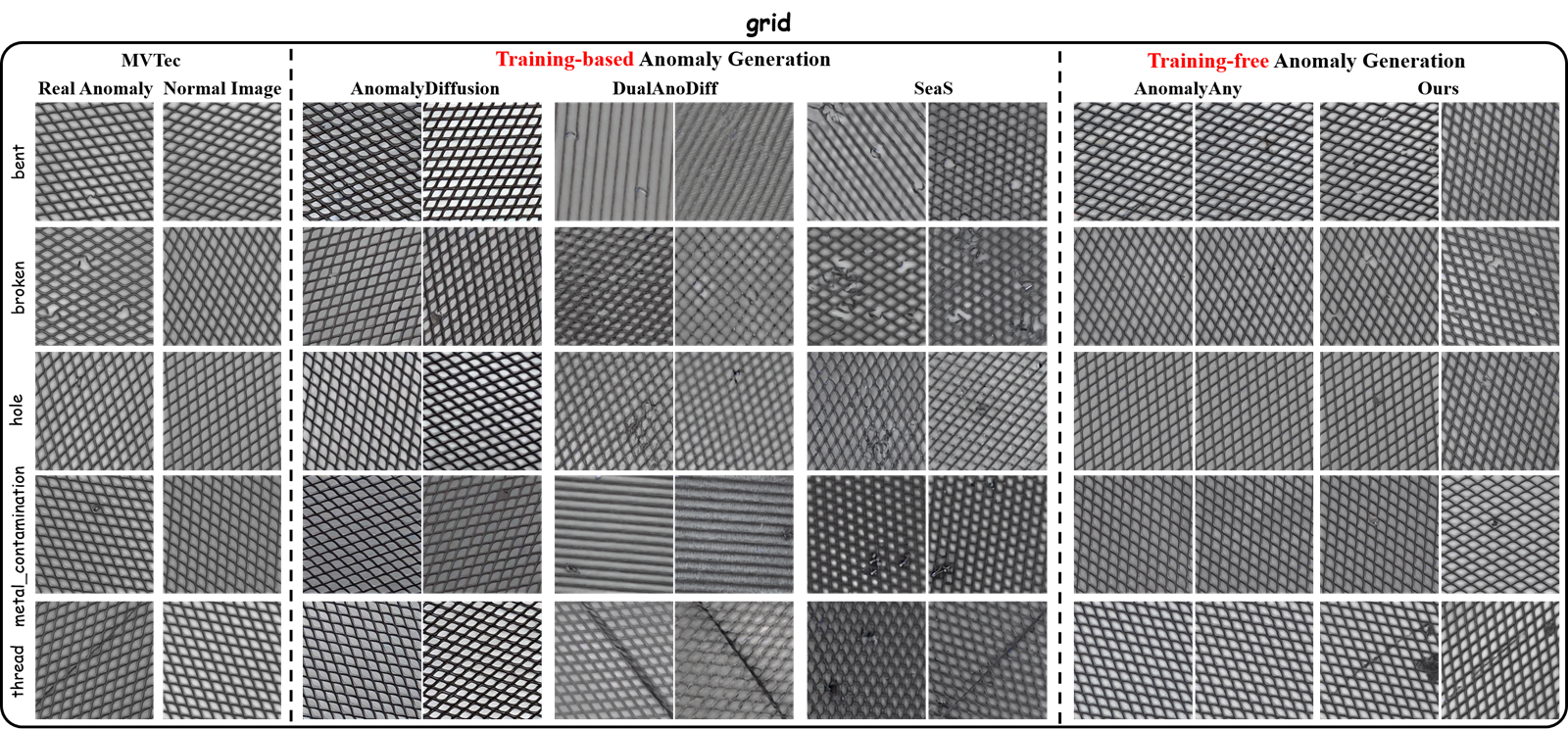}
  \caption{AnomalyDiffusion fails to synthesize the intended defects within the anomaly mask; DualAnoDiff and SeaS do not preserve background appearance and produce anomalies with distribution shift; AnomalyAny struggles to capture precise anomaly semantics. In contrast, our method preserves background fidelity and synthesizes diverse, realistic anomalies.}
  \label{fig:app-9}
\end{figure*}

\begin{figure*}[t]
  \centering
  \includegraphics[width=\textwidth]{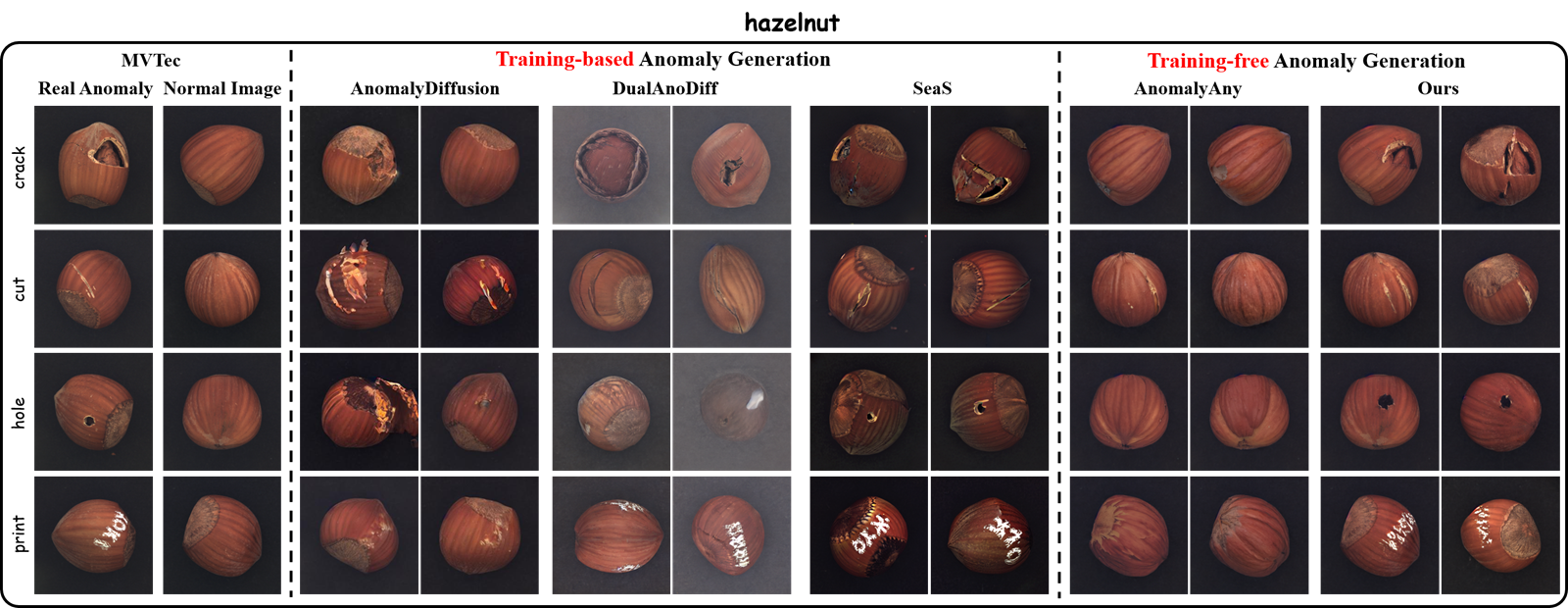}
  \caption{hazelnut qualitative results on MVTec-AD.}
  \label{fig:app-9}
\end{figure*}

\begin{figure*}[t]
  \centering
  \includegraphics[width=\textwidth]{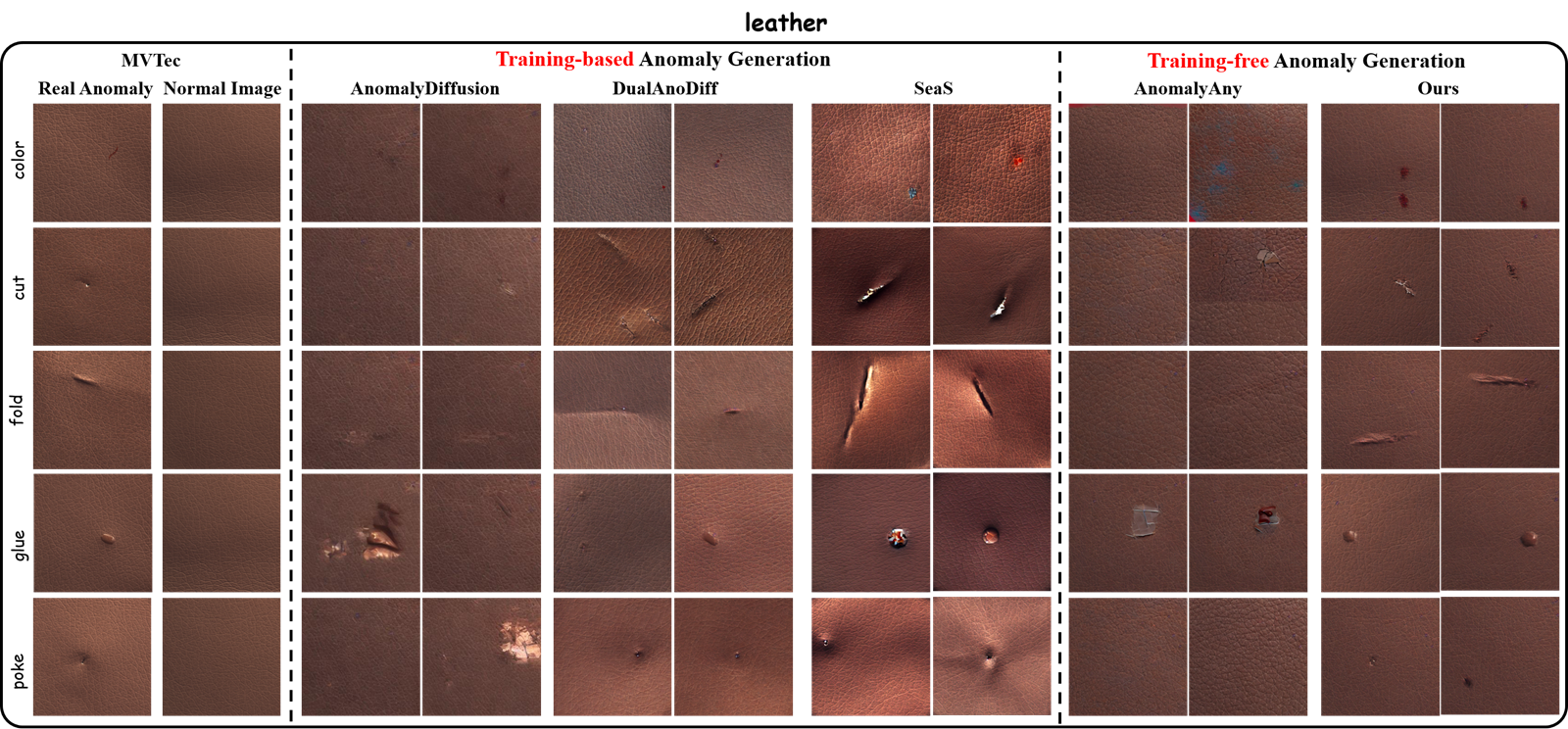}
  \caption{leather qualitative results on MVTec-AD.}
  \label{fig:app-10}
\end{figure*}

\begin{figure*}[t]
  \centering
  \includegraphics[width=\textwidth]{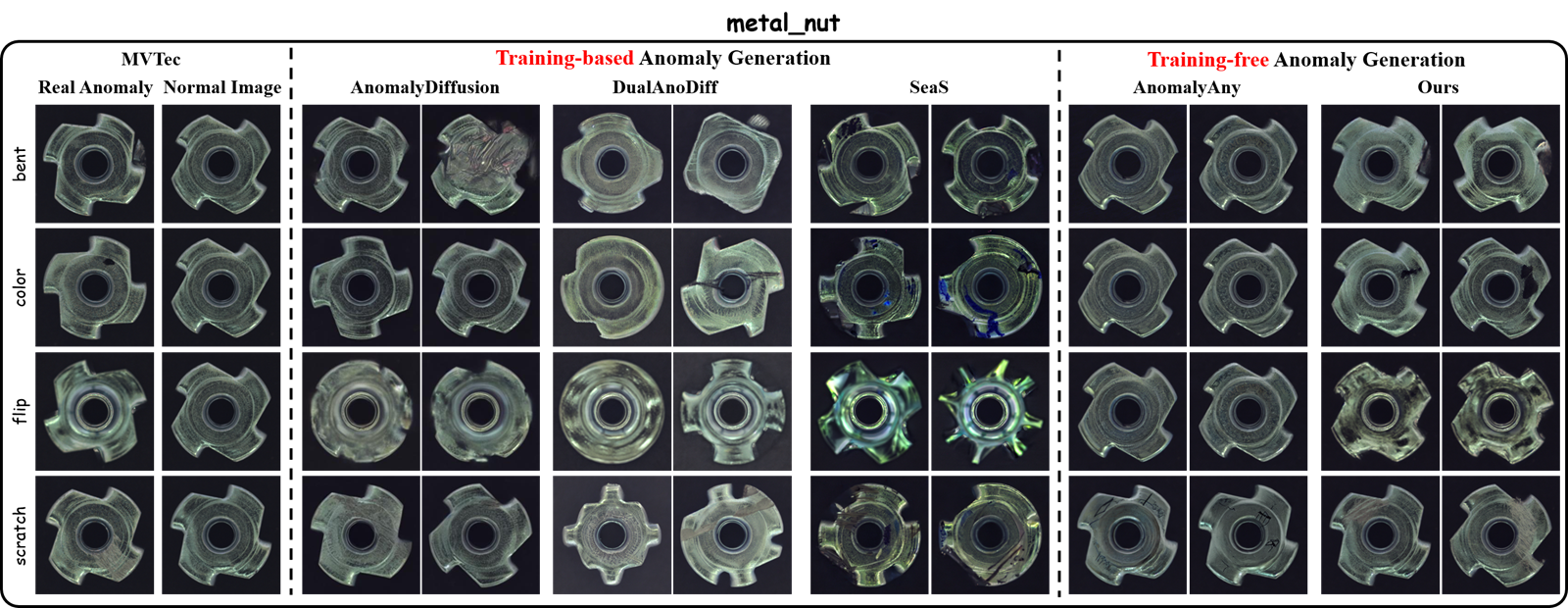}
  \caption{metal\_nut qualitative results on MVTec-AD.}
  \label{fig:app-11}
\end{figure*}

\begin{figure*}[t]
  \centering
  \includegraphics[width=\textwidth]{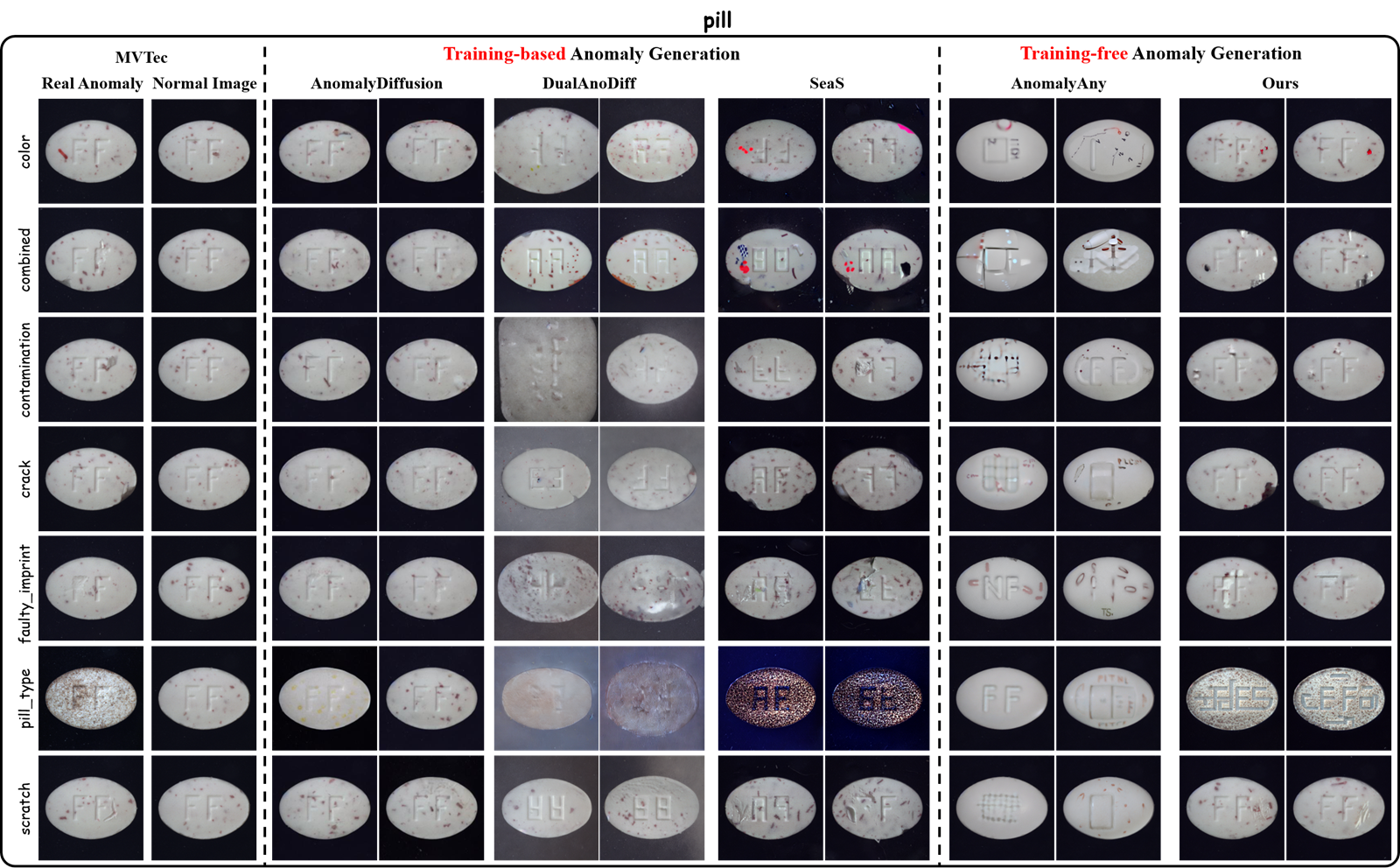}
  \caption{pill qualitative results on MVTec-AD.}
  \label{fig:app-12}
\end{figure*}

\begin{figure*}[t]
  \centering
  \includegraphics[width=\textwidth]{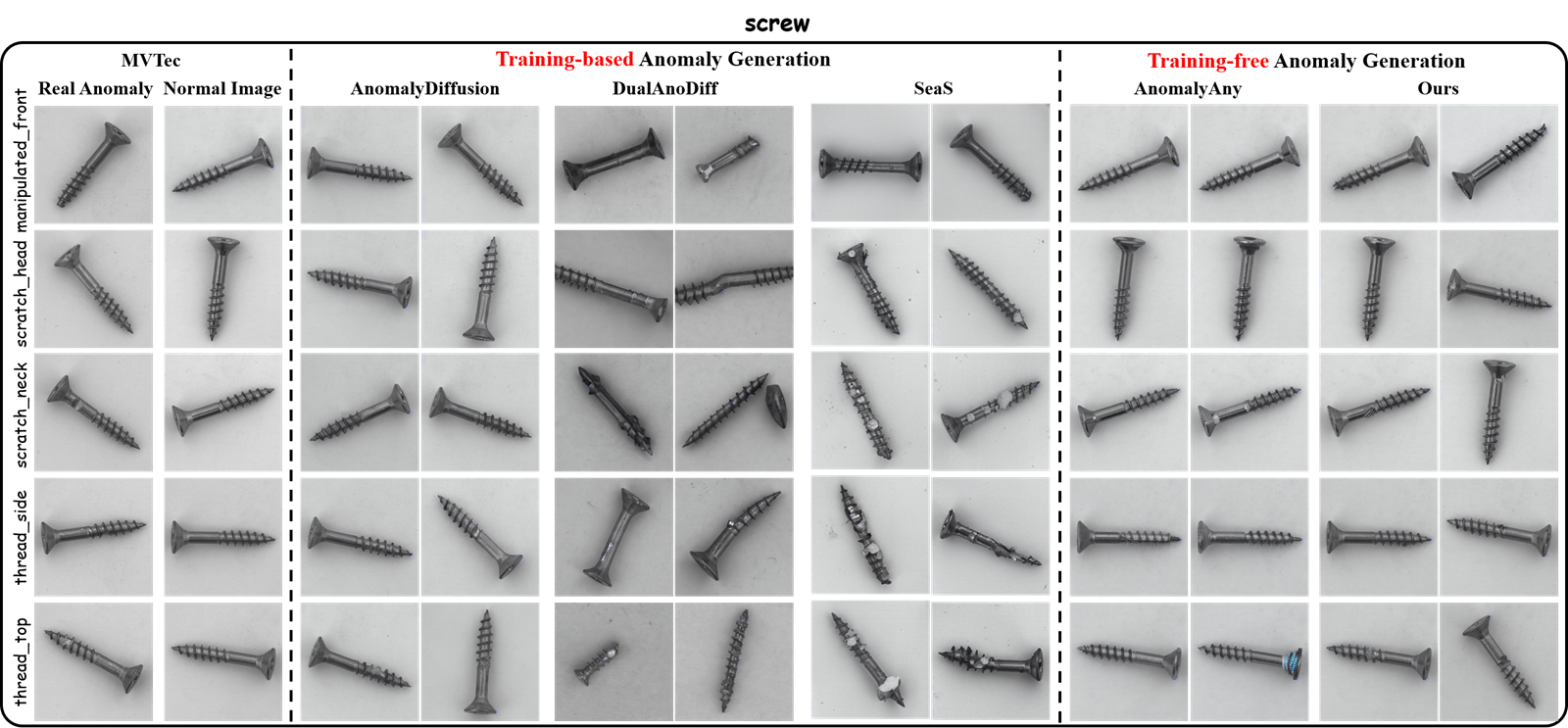}
  \caption{screw qualitative results on MVTec-AD.}
  \label{fig:app-13}
\end{figure*}

\begin{figure*}[t]
  \centering
  \includegraphics[width=\textwidth]{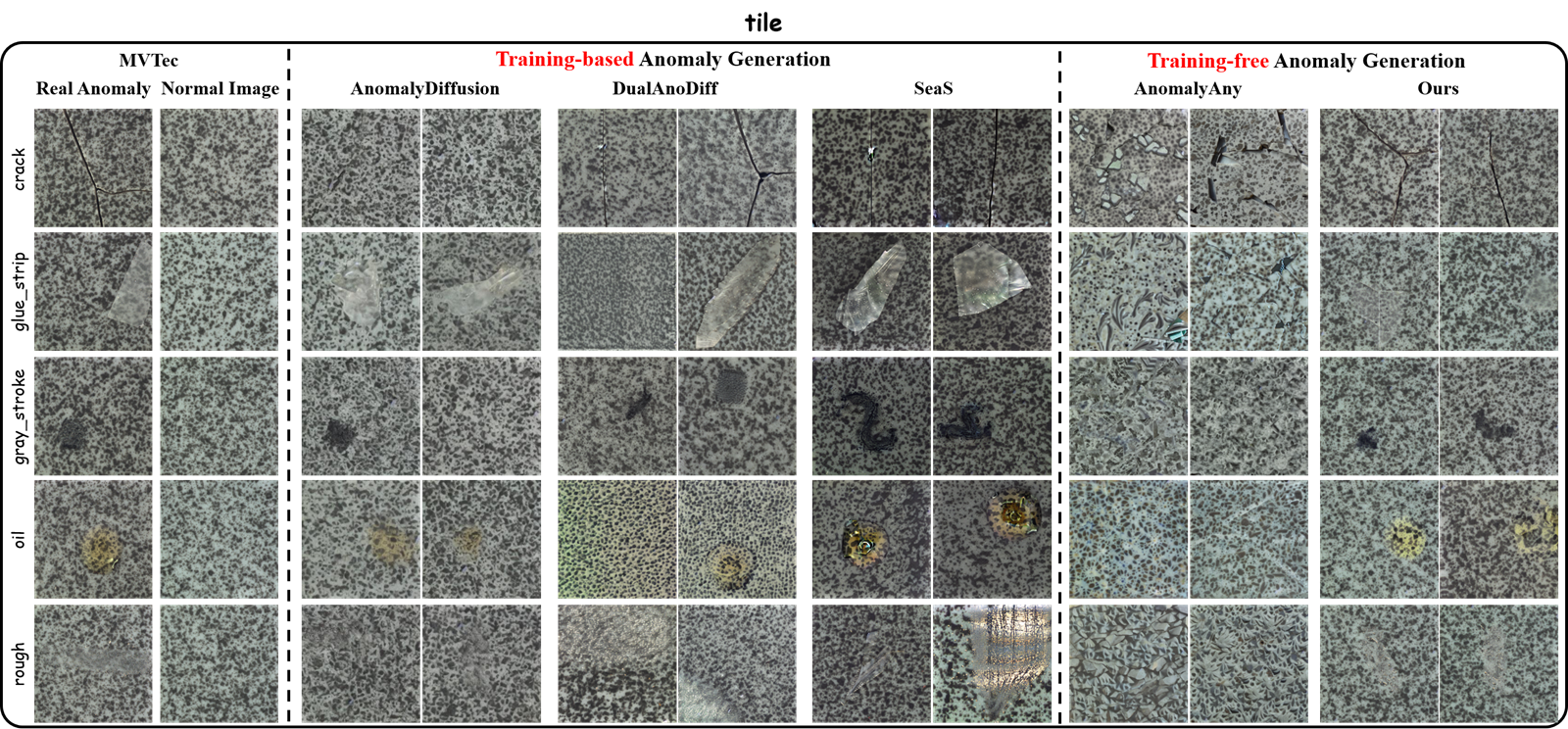}
  \caption{tile qualitative results on MVTec-AD.}
  \label{fig:app-14}
\end{figure*}

\begin{figure*}[t]
  \centering
  \includegraphics[width=\textwidth]{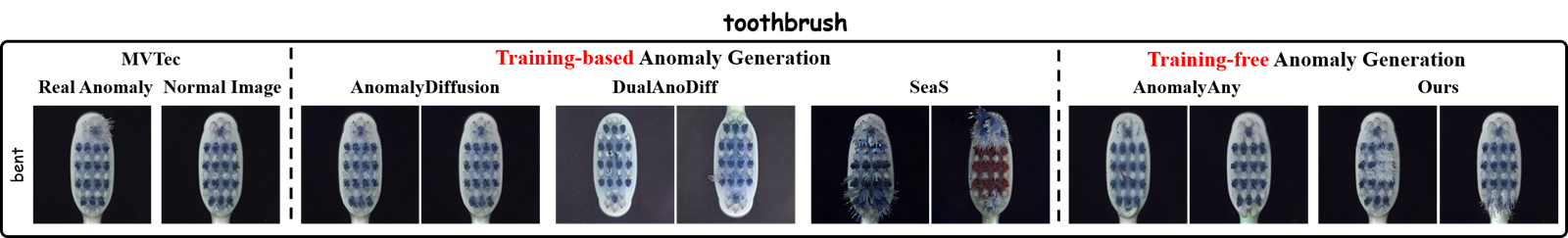}
  \caption{toothbrush qualitative results on MVTec-AD.}
  \label{fig:app-15}
\end{figure*}

\begin{figure*}[t]
  \centering
  \includegraphics[width=\textwidth]{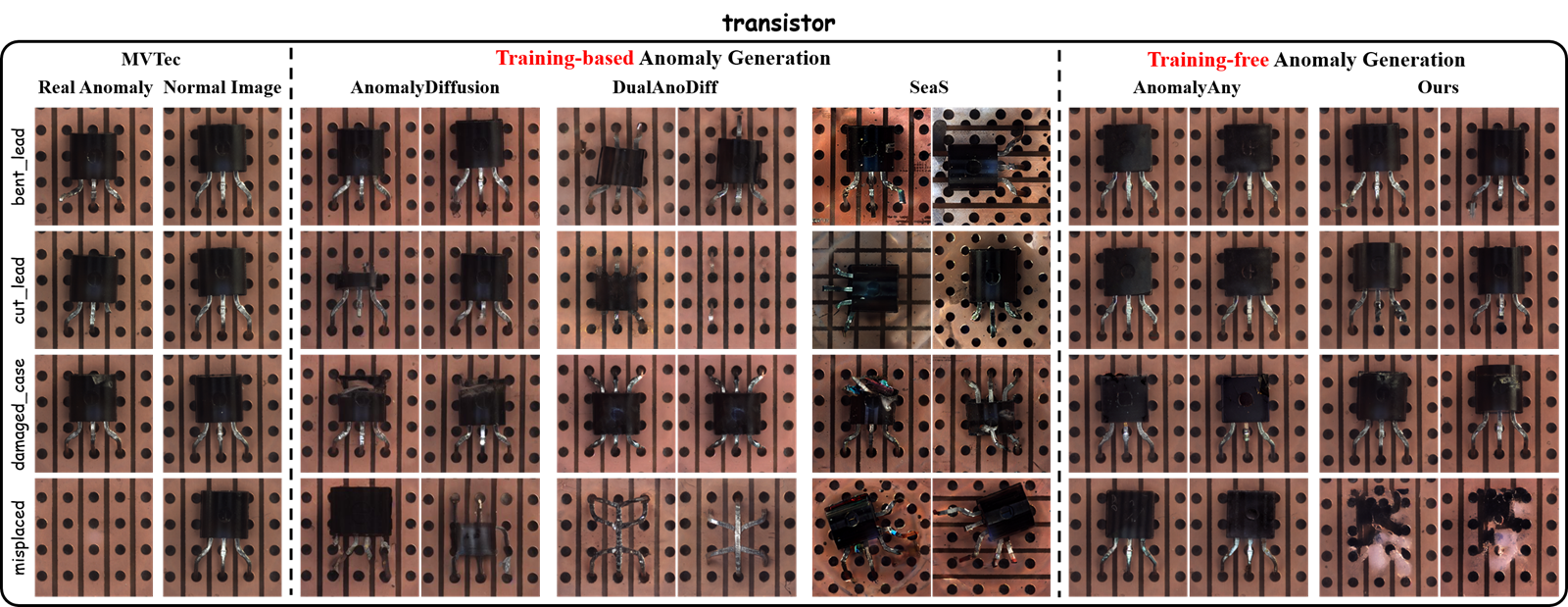}
  \caption{transistor qualitative results on MVTec-AD.}
  \label{fig:app-16}
\end{figure*}

\begin{figure*}[t]
  \centering
  \includegraphics[width=\textwidth]{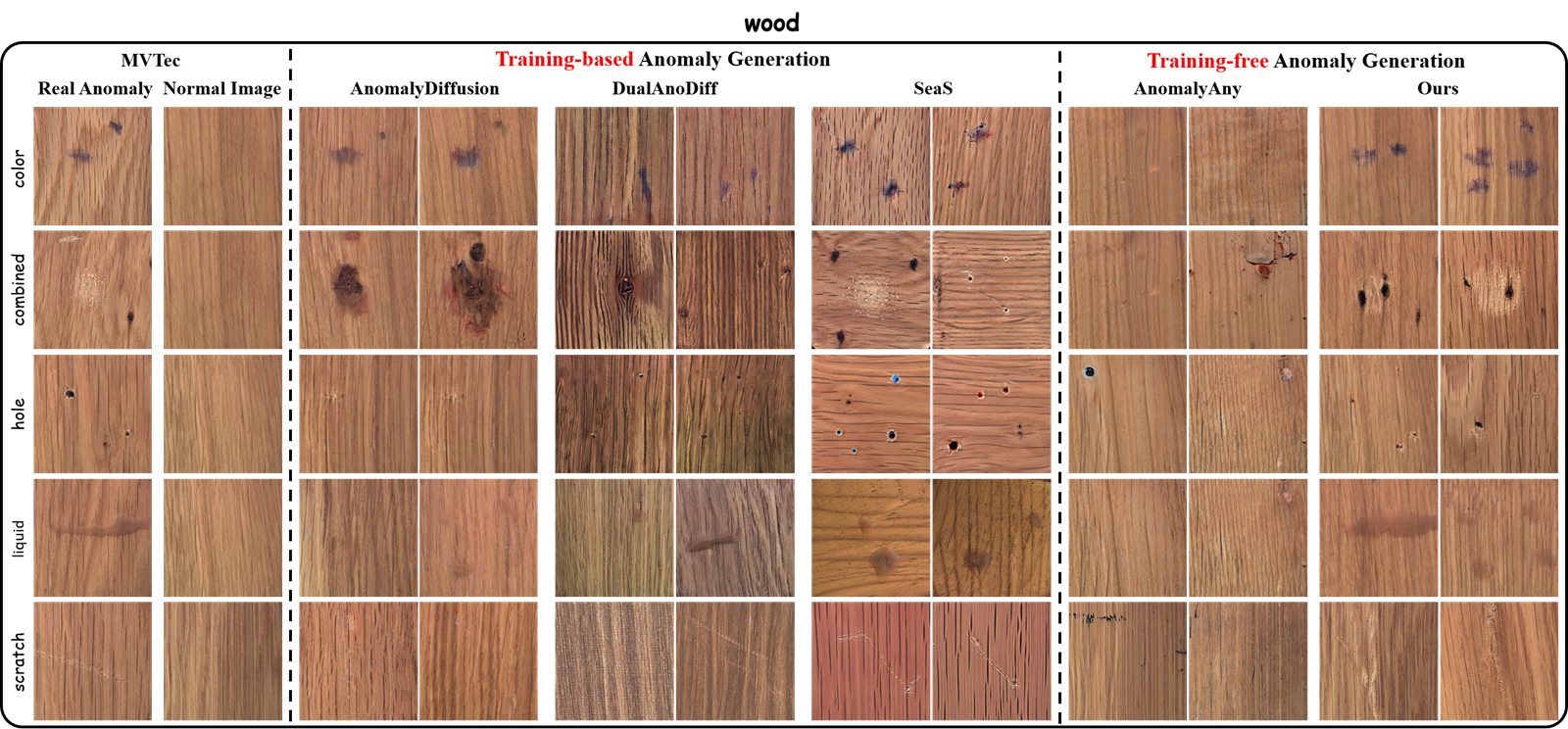}
  \caption{wood qualitative results on MVTec-AD.}
  \label{fig:app-17}
\end{figure*}

\begin{figure*}[t]
  \centering
  \includegraphics[width=\textwidth]{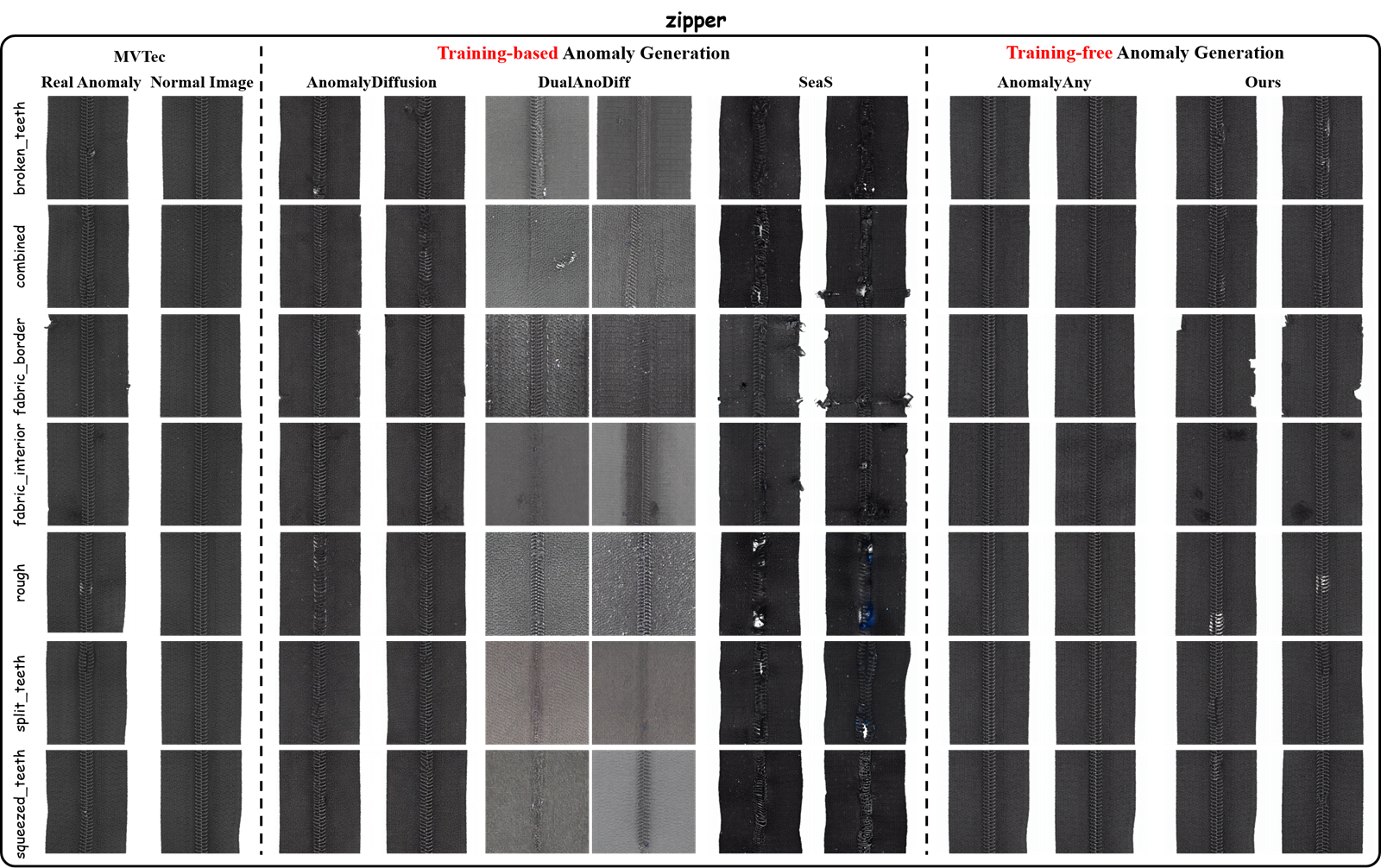}
  \caption{zipper qualitative results on MVTec-AD.}
  \label{fig:app-18}
\end{figure*}

\end{document}